\documentclass{article}

\usepackage{microtype}
\usepackage{graphicx}
\usepackage{subcaption}
\usepackage{booktabs} 
\usepackage{microtype}
\usepackage{multirow}
\usepackage{multicol}
\usepackage{graphicx}
 \usepackage{amsmath}
 \usepackage{amsfonts}
 \usepackage{booktabs}
\usepackage{subcaption}
\usepackage{booktabs} 
\usepackage{algorithm,algorithmic}
\usepackage{hyperref}

\usepackage[preprint]{icml2026}

\usepackage{amsmath}
\usepackage{amssymb}
\usepackage{mathtools}
\usepackage{amsthm}

\usepackage[capitalize,noabbrev]{cleveref}

\theoremstyle{plain}
\newtheorem{theorem}{Theorem}[section]
\newtheorem{proposition}[theorem]{Proposition}
\newtheorem{lemma}[theorem]{Lemma}
\newtheorem{corollary}[theorem]{Corollary}
\theoremstyle{definition}
\newtheorem{definition}[theorem]{Definition}
\newtheorem{assumption}[theorem]{Assumption}
\theoremstyle{remark}
\newtheorem{remark}[theorem]{Remark}

\usepackage{comment}
\usepackage{url}
\usepackage{graphicx}
\usepackage{booktabs}  
\usepackage{multirow}
\usepackage{array}         
\usepackage{xcolor}        
\usepackage{colortbl}      
\usepackage{pifont}        
\usepackage{graphicx}      
\usepackage{makecell}      

\icmltitlerunning{A Decomposable Forward Process in Diffusion Models for Time-Series Forecasting}

\begin{document}

\twocolumn[
  \icmltitle{A Decomposable Forward Process in Diffusion Models for Time-Series Forecasting }



  \icmlsetsymbol{equal}{*}

  \begin{icmlauthorlist}
    \icmlauthor{Francisco Caldas}{yyy}
    \icmlauthor{Sahil Kumar}{yyy}
    \icmlauthor{Cláudia Soares}{yyy}
  \end{icmlauthorlist}

  \icmlaffiliation{yyy}{Department of Informatics, NOVA University of Lisbon, Caparica, Portugal}

  \icmlcorrespondingauthor{Francisco Caldas}{francisco.caldas@campus.fct.unl.pt}

  \icmlkeywords{Machine Learning, ICML, Diffusion Process, Decomposition, Time-series, Fourier, Wavelet, Forecasting}

  \vskip 0.3in
]



\printAffiliationsAndNotice{}  

\begin{abstract}
We introduce a model-agnostic forward diffusion process for time-series forecasting that decomposes signals into spectral components, preserving structured temporal patterns such as seasonality more effectively than standard diffusion. 
Unlike prior work that modifies the network architecture or diffuses directly in the frequency domain, our proposed method alters only the diffusion process itself, making it compatible with existing diffusion backbones (e.g., DiffWave, TimeGrad, CSDI).
By staging noise injection according to component energy, it maintains high signal-to-noise ratios for dominant frequencies throughout the diffusion trajectory, thereby improving the recoverability of long-term patterns.
This strategy enables the model to maintain the signal structure for a longer period in the forward process, leading to improved forecast quality. 
Across standard forecasting benchmarks, we show that applying spectral decomposition strategies, such as the Fourier or Wavelet transform, consistently improves upon diffusion models using the baseline forward process, with negligible computational overhead.
The code for this paper is available at \url{https://anonymous.4open.science/r/D-FDP-4A29}.

\end{abstract}

\section{Introduction}

Time-series forecasting plays a critical role in a wide range of real-world applications~\citep{Bryan2021}, from weather prediction~\citep{Choi2023} to financial market analysis~\citep{MAKRIDAKIS2024}. Classical statistical methods~\citep{Box2016} and autoregressive linear models~\citep{hyndman} remain widely used due to their interpretability and competitive predictive performance. However, these approaches often struggle with capturing long-range dependencies~\citep{Zhou_Zhang_Peng_Zhang_Li_Xiong_Zhang_2021}, hierarchical seasonal structures~\citep{hewamalage2021recurrent}, and non-linear dynamics~\citep{rangapuram}. 
Diffusion models~\citep{Sohl-Dickstein_2015}, which learn a Markov chain of progressively denoised latent representations, have demonstrated state-of-the-art performance in image generation~\citep{Ho_2020,Dhariwal_2021} and inpainting~\citep{Lugmayr_2022_CVPR}. More recently, they have been applied to time-series forecasting~\citep{Kong_2020,Tashiro_2021}, showing competitive results against other generative models.  While some works have incorporated multi-resolution processing~\citep{shen2024multiresolution,Jiatao_2022,fan2024mgtsd}, these models do not fully exploit the inherent structure of time-series data in their diffusion processes. 

A well-established technique in time-series modeling is decomposition into three main components: trend, seasonal effects, and residual noise~\citep{hyndman,cleveland90,yi2023frequencydomain}. However, existing diffusion models lack mechanisms to preserve these structured components, often leading to inaccurate long-term predictions. For instance, in electricity demand forecasting, standard diffusion-based models tend to smooth out periodic consumption cycles, failing to accurately capture sub-daily patterns. This loss of structured information diminishes the model’s forecasting ability, especially in applications that require precise seasonality preservation. 

To address this limitation, we introduce a novel approach that integrates structured time-series decomposition into the forward diffusion process. Specifically, we employ spectral decomposition to isolate and process seasonal components and remainders separately. After identifying the different components, our proposed diffusion process performs diffusion sequentially and in ascending order of amplitude, guaranteeing that the more relevant components are kept longer in the forward diffusion process and reconstructed earlier. 
Our method modifies the sample generation of the training process, and equivalently the sample inference process, which means that it can be used with any known deep learning architecture for time-series model. This approach is particularly suited for long time-series, where stronger seasonal patterns are common and can be more effectively captured through seasonal decomposition during both training and inference, in the conditional window.

Furthermore, we evaluate this framework using different types of spectral decomposition and show that it is generalizable to any additive decomposition method. We demonstrate its efficacy using both Fourier and wavelet transforms, and show consistent improvements over diffusion-based forecasting models that employ the conventional forward process.

\textbf{Contributions.}
Existing diffusion-based time series models (e.g., DiffWave~\citep{Kong_2020}, TimeGrad~\citep{Rasul_2021}, MG-TSD~\citep{fan2024mgtsd}, mrDiff~\citep{shen2024multiresolution}, Diffusion-TS~\cite{yuan2024diffusionts}) focus primarily on adapting architectural components, such as attention or S4 layers \cite{gu2022s4}, or developing hierarchical models, but retain the standard forward process that injects noise indiscriminately, erasing structured temporal features such as seasonality and trend. 
In contrast, we redesign the forward process itself to explicitly preserve these features, yielding more faithful reconstructions and adapted to long-horizon forecasting 
when structure-aware extrapolation is required. Unlike previous works, our approach is model agnostic and can be applied to any diffusion model architecture.

Our contributions are threefold: (1) we propose a novel structured forward diffusion process leveraging spectral decomposition for time-series forecasting; (2) we theoretically generalize this framework to accommodate various additive decomposition methods; and (3) we empirically validate our spectral decomposition approach on multiple benchmark datasets, demonstrating its effectiveness in capturing long-range dependencies and seasonal patterns.

\begin{figure*}[t]
    \centering
    \includegraphics[width=\textwidth]{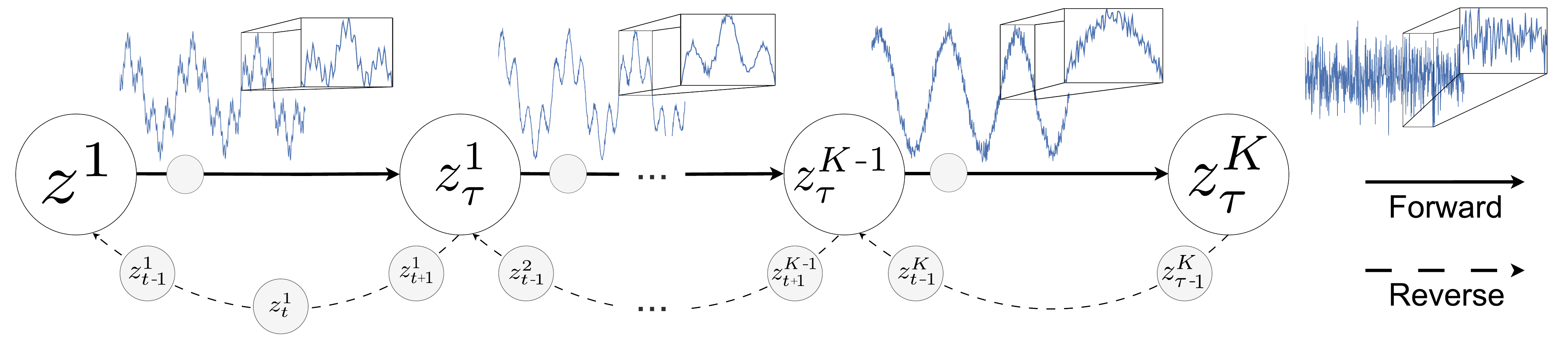}
    \caption{Diffusion process with decomposition. During the forward process, noise is added to each component in ascending order of amplitude. This process slows down the destruction of the more relevant frequencies, preserving key structures longer.}
    \label{fig:diffusion_1}
\end{figure*}

\section{Related Work}
\label{sec:back}

\paragraph*{i. Diffusion Models}
Diffusion probabilistic models~\citep{Sohl-Dickstein_2015,Song_2020} have gained significant popularity since their introduction, demonstrating remarkable success in image generation~\citep{Ho_2020,Ho_2021} and text-to-speech synthesis~\citep{natural}. They have also been extended to video \citep{ho2022video}, audio~\citep{Chen_2020}, and other applications \citep{luo2021diffusion}. \citet{Song_2020} unified two perspectives on these models—discrete and continuous formulations—under the umbrella of score-based generative models. In this work, we adopt the discrete formulation, equivalent to the variance-preserving stochastic differential equation (VP-SDE)~\cite{Ho_2020}, in a non-autoregressive setting.

\paragraph*{ii. Latent and Multi-scale Diffusion} Recent advancements have improved diffusion models in terms of scalability and efficiency. Techniques such as inference cost reduction~\citep{Song_2020}, and others~\citep{song2020improved,bortoli2021diffusion,watson2021learning} have enhanced their applicability. Additionally, \citet{Nichol_2021} introduced variance prediction in the reverse diffusion process, enabling competitive performance with fewer sampling steps. For high-resolution image generation, latent diffusion models~\citep{Ramesh_2022} compress data into a lower-dimensional space before applying the diffusion process. A different approach applies diffusion in the spectral domain \citep{phillips2022spectral,crabbetime}.  Other hierarchical models refine outputs by generating a low-resolution image first, followed by super-resolution techniques~\citep{Jiatao_2022,Ho_2021}. However, these improvements remain largely focused on spatial data, and their direct applicability to time-series forecasting remains limited due to differences in structural dependencies.



%

%

\paragraph*{iii. Temporal Representation}
Recent research has focused on improving temporal representations in Transformer-based forecasting models. Fedformer~\citep{zhou2022fedformer} maps time-series data into the frequency domain, selecting a random subspace for modeling, while Pyraformer~\citep{liu2022pyraformer} introduces pyramidal attention to enable multi-resolution analysis. Basis expansion techniques, such as N-Beats~\citep{Oreshkin2020:N-BEATS}, decompose time-series data into trend and seasonal components, with subsequent refinements like N-Hits~\citep{challu2023nhits},~\citet{zeng} and \citet{fons2024ihypertime} improving multi-scale modeling and simplifying decomposable time-series structures into linear models. 
In addition, frequency-aware models like \citet{wu2023timesnet,cyclenet}, and \citet{wang2023timemixer,wang2024timemixer++} have presented new layers that are specifically tailored to enhance the ability to extract temporal and multi-seasonal information using Fourier Decomposition to extract top-$k$ frequencies. The first TimeMixer paper \citep{wang2023timemixer} hinted during ablation studies that FFT-based decomposition yielded better results than moving-average decomposition, while the latter expanded on the use of Fourier decomposition alongside MixerBlocks and dual-axis attention.

\paragraph*{iv. Diffusion Time-series} The application of diffusion models to time-series forecasting began with WaveNet~\citep{deeOord2016} and latter DiffWave~\citep{Kong_2020}, originally designed for audio and speech generation. Building on DiffWave, TimeGrad~\citep{Rasul_2021} applied a similar architecture to diffusion-based forecasting using an auto-regressive setting and Langevin dynamics. Around the same time, ScoreGrad~\citep{Yan_2021} introduced a stochastic differential equation (SDE) formulation for time-series diffusion. Subsequent works have proposed incremental improvements, including Transformer-based architectures~\citep{Tashiro_2021}, structured state-space models~\citep{Lopez_2022}, Fourier-based loss \cite{yuan2024diffusionts}, and multi-resolution forecasting via future mixup~\citep{Shen_2023,shen2024multiresolution}. The latter employs a multi-stage approach to inference, where different models process varying levels of data granularity. However, these methods rely on fixed-stage decomposition, which can lead to information loss in highly dynamic time-series data. A similar approach \cite{wang2025a} used fixed moving averages, to define a non-diagonal transition matrix for the forward process, as a way to slowdown the diffusion process across frequency bands.
Fundamentally, none of these approaches incorporate structured time-series decomposition directly into the diffusion process, or use a fixed periodicity/kernel size, limiting their ability to preserve long-range dependencies and detected periodic patterns. 

%

\textbf{This paper.} While prior work has made progress in diffusion-based time-series forecasting, it has not addressed the fundamental issue of structured information loss in the forward diffusion process. In contrast, our approach explicitly integrates structured time-series decomposition within the diffusion framework. By leveraging Fourier decomposition, we ensure that periodic components are preserved throughout the forward process (Figure~\ref{fig:diffusion_1}), improving forecasting accuracy for seasonal and long-horizon time-series. Unlike prior multi-resolution methods, our approach is fully generalizable to different additive decomposition techniques and does not impose fixed granularity constraints.

Unlike prior approaches that operate solely through model architecture or conditioning (e.g., Transformer-based or Cascaded models), our contribution introduces a structured forward diffusion process that incorporates signal decomposition, enabling frequency-aware denoising and better temporal structure retention~(Table~\ref{tab:model_comparison_ticks}).

\begin{table*}[h]
\centering
\caption{Feature comparison of FFT-DM with prior diffusion-based time-series models.}

\label{tab:model_comparison_ticks}

\resizebox{0.99\textwidth}{!}{
\begin{tabular}{@{}p{3.5cm}ccccccc>{\columncolor{gray!15}}c@{}}
\toprule

\textbf{Feature} 

& \makecell{DiffWave \\ \cite{Kong_2020}} 
& \makecell{TS-Diffusion \\ \cite{yuan2024diffusionts}} 
& \makecell{TimeGrad \\ \cite{Rasul_2021}} 
& \makecell{CSDI \\ \cite{Tashiro_2021}} 
& \makecell{MG-TSD \\ \cite{fan2024mgtsd}} 
& \makecell{mrDiff \\ \cite{shen2024multiresolution}}
& \makecell{MA-TSD \\ \cite{wang2025a}}
& \textbf{\makecell{Decomposable\\ Forward Process (Ours)}} 
\\ \midrule
Decomposition modules    &  &  &  &  & \checkmark & \checkmark &  & \checkmark \\
Frequency awareness            &  & \checkmark &  &  & \checkmark & \checkmark & \checkmark &  \checkmark \\
Architecture agnostic          &  &  &  &  &  &  & \checkmark &\checkmark \\
Non-autoregressive  & \checkmark & \checkmark &  & \checkmark &  & \checkmark & \checkmark& \checkmark \\
Seasonal/trend structure       &  & \checkmark &  &  & \checkmark & \checkmark & & \checkmark \\
Forecasting                    &  &  & \checkmark & \checkmark & \checkmark & \checkmark & \checkmark& \checkmark \\
Flexible Decomposition & & & & & & & & \checkmark \\
\bottomrule
\end{tabular}
}
\end{table*}


\section{Diffusion Models}

\paragraph*{Diffusion Forward Process.}
A diffusion process~\citep{Sohl-Dickstein_2015} consists of progressively adding Gaussian noise to a sample $x_0$ over $T$ time-steps, such that the data distribution transitions to an isotropic Gaussian distribution. The process state transition is defined as
\begin{equation}
    q(x_t|x_{t-1}) = \mathcal{N}(x_t|\sqrt{1-\beta_t}x_{t-1},\beta_t\mathbf{I}),
    \label{eq:forw}
\end{equation}
or, equivalently,
\begin{equation}
    x_t = \sqrt{1-\beta_t}x_{t-1} + \sqrt{\beta_t}\epsilon,
    \label{eq:forw2}
\end{equation}
where $\epsilon \sim \mathcal{N}(0, \mathbf{I})$, and $\beta_t$ is a noise scheduling parameter controlling the rate of information loss. The values of $\beta_t \in ]0,1[$ are defined according to a scheduler, such as a linear or cosine schedule~\citep{Ho_2020}, and follow the constraint $\beta_1 < \beta_2 < \dots < \beta_T$. 

Given independent noise at each time-step $t$, the closed-form expression for $x_t$ at any arbitrary $t$ is
\begin{equation}
    x_t = \sqrt{\bar{\alpha_t}}x_0 +\sqrt{1-\bar{\alpha_t}}\epsilon,
    \label{eq:close}
\end{equation}
where $\alpha_t = (1-\beta_t)$ and $\bar{\alpha}_t = \prod_{i=1}^{t} \alpha_i$ and $q(x_{1:t}|x_0) = \prod_{i=1}^t q(x_i|x_{i-1})$. As $T \to \infty$, the data distribution converges to a Gaussian prior
\begin{equation}
    q(x_T|x_0) = q(x_T) = \mathcal{N}(0, \mathbf{I}),
\end{equation}
that is independent of $x_0$.

This formulation defines the forward diffusion process, which is the focus of this work. In the following sections, we adapt these equations to the case where $x$ has a known lossless decomposition.

\paragraph*{Reverse Diffusion Process.}

The goal of diffusion models is to learn a denoising function that approximates the reverse process $q(x_{t-1}|x_t)$. Since this transition is intractable~\citep{Bishop_2024}, an approximation is used:
\begin{equation}
    p(x_{t-1}|x_t, \Theta) = \mathcal{N}(x_{t-1} | \boldsymbol{\mu}_\theta(x_t, t), \sigma_t \mathbf{I}),
\end{equation}
where $\Theta$ represents the parameters of the neural network, which learns to estimate the mean function $\mu_\theta(x_t, t)$. Under the assumption that $\beta_t \ll 1$, the true posterior $q(x_{t-1}|x_t, x_0)$ is well-approximated by a Gaussian distribution~\citep{Sohl-Dickstein_2015}. The variance term $\sigma_t$ is often assumed to be known and fixed, although some models also learn it as part of the diffusion process~\citep{Nichol_2021}.



\paragraph*{Training Objective.}

The model is typically trained by minimizing the Evidence Lower BOund (ELBO)~\citep{KingmaW13Elbo}, which can be formulated in multiple ways: as a data prediction task, a noise estimation problem, or a step-wise reconstruction objective. The simplified objective is derived from the sum of Kullback-Leibler (KL) divergences:
%
\begin{align}
    \mathcal{L} =& \sum_{t=2}^T KL(q(x_{t-1}|x_t,x_0)||p_\theta(x_{t-1}|x_t)) \\
    \mathcal{L} =& \sum_{t=2}^T \mathbb{E}_{q(x_t|x_0)}\!\left[\bigl\| m(x_t,x_0) - \boldsymbol{\mu}_\theta(x_t,t)\bigr\|^2\right] + c.
    \label{eq:solh}
\end{align}

Another approach involves rewriting $\boldsymbol{\mu}_\theta(x_t,t)$ as a function of a noise model \citep{luo2022understanding}:
\begin{equation}
    \boldsymbol{\mu}_\theta(x_t,t) = \frac{1}{\sqrt{\alpha_t}}\left(x_t - \frac{\beta_t}{\sqrt{1-\bar{\alpha_t}}} \right)\boldsymbol{\epsilon}_\theta(x_t,t).
    \label{eq:ho}
\end{equation}

In this case the noise predicting model $\mathbf{\epsilon}(x_t,\Theta,t)$ leads to the  simplified loss 

\begin{equation}
    {\mathcal{L}_{\text{simple}}} = \mathbb{E}_{t,x_0,\epsilon} \left[\|\epsilon -\boldsymbol{\epsilon}_\theta(x_t,t)\|^2\right].
\end{equation}

This formulation allows predicting the total noise added to the data using~\eqref{eq:close}, a technique successfully applied in time-series diffusion models~\citep{Lopez_2022, Tashiro_2021}.

\paragraph*{Conditional Diffusion for Time-Series Forecasting.}

In time-series forecasting, a special case of imputation, and the focus of this work, the goal is to predict future values $\mathbf{x}_{0}^{P+1:W} \in \mathbb{R}^{d \times (W-P)}$ given past observations $\mathbf{x}^{0:P}_0 \in \mathbb{R}^{d \times P}$. Here, $W-P$ is the length of the forecast window, and $P$ is the length of the lookback window. In the most general case, let $\mathbf{x}_{0}^{Obs}$ denote the observed time-steps and $\mathbf{x}_{0}^{Tgt}$ the target time-steps.

%
In conditional diffusion models, the denoising process at step $t$ is given by~\citep{shen2024multiresolution}:
%
%
\begin{align}
 p_\theta(\mathbf{x}_{t-1}^{Tgt}|\mathbf{x}_{t}^{Tgt}, \mathbf{c}) = \mathcal{N}(\mathbf{x}_{t-1}^{Tgt}; \mu_\theta(\mathbf{x}_{t}^{Tgt}, \mathbf{c}, t), \beta_t\mathbf{I}),
%
\end{align}
where $\mathbf{c} = \mathcal{F}(\mathbf{x}_{0}^{Obs}, t)$ represents the conditioning information extracted from past observations. The transformation $\mathcal{F}$ encodes relevant features such as past time-series values and temporal embeddings. For simplicity, in the remainder of this paper, we omit notation explicitly indicating this conditioning when it is clear that the model is conditioned on all past time-steps.


\section{Decomposable Diffusion Model}
\label{sec:al}

\paragraph*{Forward Process.}


We propose a decomposable Forward Diffusion Process, a model-agnostic modification of the diffusion forward process for time-series forecasting. In standard diffusion models, noise is applied indiscriminately to the entire input~\citep{Ho_2021}, so every diffusion step simultaneously corrupts all aspects of the signal. In contrast, many time-series can be decomposed into disjoint spectral components (e.g., seasonalities and residuals) whose sum reconstructs the original signal. Our process leverages this additive property by staging noise injection across components, so that the reverse diffusion process focuses on reconstructing the dominant structures first before addressing finer residuals. This sequential decomposition does not add learning complexity, preserves periodic patterns longer into the diffusion trajectory, and can be applied to any backbone without architectural changes.

To formalize this, we introduce a {\bf generalized forward diffusion process} that incorporates signal decomposition and provides a closed-form expression for each forward step at time $t$. First consider that a sample $x_0$ is expressed as the sum of its decomposed components:
\begin{equation}
    x_0 = z_0 = \sum_{k=1}^{K} f^k_0
\end{equation} 

where $D$ is the operator that linearly decomposes the sample $z_0$ into $K$ components $\{f_i^0\}^K_{i=1} = D(z_0)$. These components are, by definition, orthogonal.

Considering that $\{ z_t^k \}_{t = 1, \dots, \tau}^{k = 1, \dots, K}$, is the set of latent variables that will be transformed from the initial data $z_0$ to a gaussian distribution, the initial step $q(z_1^1|z_0)$ is defined as:
\begin{equation}
    z_1^1 = \sqrt{1-\beta_1}f^1_0  + \sqrt{d_1\beta_1}\epsilon + \sum_{k=2}^{K} f^k
\end{equation}

where $\{\beta_i\}^\tau_{i=1}$, is the noise scheduling parameter of the staged diffusion, from $i$ to $\tau$, where $\tau$ is the period of each staged, such that $\tau K = T$, and $d_i$ is a scaling factor.
Note that, since the staged diffusion runs only for $\tau \leq T$ steps, 
the noise schedule parameters $\{\beta\}_{t=1}^\tau$ and their cumulative products $\bar{\alpha}_t = \prod_{t=1}^i (1-\beta_t)$ 
are defined with respect to this shortened horizon.
The end of the first stage is $q(z^1_\tau|z^1_{\tau-1})$:
\begin{align}
     z^1_{\tau} &= \sqrt{1-\beta_\tau}f_{\tau-1}^1  + \sqrt{d_1\beta_\tau}\epsilon + \sum_{k=2}^{K} f^k \\
     z^1_\tau &= f^1_\tau + \sum_{k=2}^{K} f^k = z^2_0
     \label{complete}
\end{align}
and $q(z^2_1|z_0^2)$:
\begin{equation}
    z^2_1 = \sqrt{1-\beta_1}f_{0}^2  + \sqrt{d_2\beta_1}\epsilon + \sum_{k=3}^{K} f^k + f^1_\tau
\end{equation}

Here, $K$ represents the number of significant components, and always contains a component that corresponds to the residual term.  This ensures a {\bf lossless decomposition}, and trivially the usual standard diffusion process occurs when $K=1$. 

\begin{figure*}[!t]
    \centering
    \includegraphics[width=0.99\textwidth]{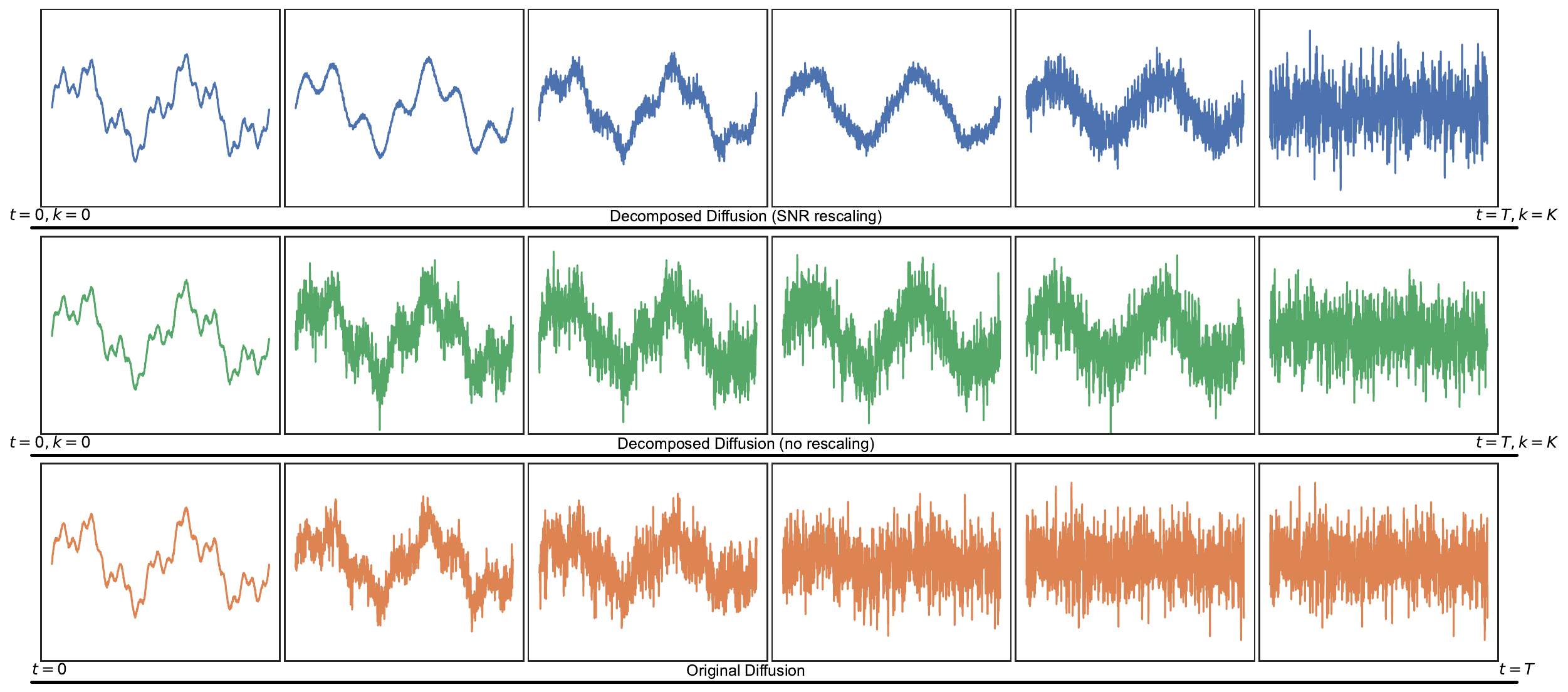}
    \caption{Effect of SNR scaling in noise application. Key frequencies in the signal remain discernible for a longer duration. Without SNR scaling, noise spreads across components, degrading separability. All approaches in this figure use a linear schedule $0.002 < \beta < 0.02$.}
    \label{fig:diffusion_2}
\end{figure*}

To regulate noise diffusion, we introduce a {\bf signal-to-noise ratio (SNR) scaling factor} $d_k$~\citep{oppenheimsignais}. This factor adjusts the noise added to each frequency component, preventing excessive diffusion in the early stages and ensuring that noise does not spill over into subsequent components (Figure~\ref{fig:diffusion_2}). Other works have empirically \cite{wang2025a} and theoretically shown that the usual DDPM approach leads to fast suppression of high-frequency signals \cite{chen2025singlestep}, and the steep replacement of low frequency information with high-frequency noise. Following these works, our goal is to delay the destruction of information by the diffusion process and to guarantee that this more controlled diffusion process occurs in the appropriate component step.

In Appendix \ref{sec:SNR} we evaluate in detail the SNR of our proposed diffusion process. This idea has also been explored in hierarchical and cascading diffusion models~\citep{Gu_2023, Jiatao_2022}.



Applying ~\eqref{eq:forw2} to $f^k$ with the SNR, meaning that when $t =\tau$, $f^k_\tau \sim N(0,d_kI)$ we derive:

\begin{equation}
    f^k_{t} = \sqrt{1-\beta_t} f^k_{t-1} + \sqrt{d_k \beta_t} \epsilon,
    \label{eq:firstforward}
\end{equation}
and 
\begin{equation}
    z^k_t = f^k_t + \sum_{n>k}^Kf_0^n + \sqrt{\sum_{n=1}^{k-1} d_n}\epsilon.
\end{equation}

This leads to the closed-form expression for $q(z_t^k | z_0)$:
\begin{align}
z_t^k &= \sqrt{\bar{\alpha}_t} f_0^k + \sum_{n>k}^{K} f_0^n + \sqrt{d_k(1-\bar{\alpha}_t)} \epsilon + \sqrt{\sum_{n=1}^{k-1} d_n} \epsilon   \notag \\
z_t^k &= \sqrt{\bar{\alpha}_t} f_0^k + \sum_{n>k}^{K} f_0^n + \sqrt{-d_k \bar{\alpha}_t + \sum_{n=1}^k d_n} \epsilon,
\label{eq:formula}
\end{align}
for diffusion step $t$ in stage $k$, with $\epsilon \sim N(0,I)$. The term $\sum_{n>k}^{K} f_0^n$ represents the frequencies/components yet to be diffused, while the frequencies already diffused are defined as  $\sum_{n=1}^{k-1}d_k$, in the first line of~\eqref{eq:formula}, using the property of the variance of the sum of independent variables. note also that, as can be inferred from \eqref{complete}, $z_\tau^k = z_0^{k+1}$ for $k<K$ and $z_\tau^K \sim N(0,I)$

The {\bf SNR scaling factor} ($d_k)$ for each component is computed as: \begin{equation} \text{SNR} = \frac{\mathbb{E}[(f^k)^2]}{\mathbb{E}[\epsilon^2]}, \end{equation} ensuring that noise diffusion is proportional to the amplitude of each frequency component. We highlight that this scaling factor is calculated based on the components, and therefore is not a hyperparameter. In Fig. \ref{fig:diffusion_2} we empirically show that SNR rescaling is necessary to guarantee that each component is diffused during its respective period. In particular, we use Fourier Decomposition and Wavelet decomposition to validate our approach, explained in detail in the Appendixes \ref{fourier} and \ref{wavelet}, however, the diffusion process itself remains {\bf agnostic to the specific decomposition method}, as long as it satisfies the weak assumption of lossless reconstruction.

\paragraph*{Reverse Process.}

As we are applying a sample specific decomposition with Signal-to-noise-ratio $d_k$, which might not be constant, during the inference process this value is unknown. One possible solution is to consider $d_k$ as part of the output of the model, modifying the Loss function in~\eqref{eq:ho}. However, in this case, we estimate these values from the conditioning time-steps during inference. This is equivalent to performing the decomposition on the masked samples, however in this case we are only interested on the amplitude of the component, and not the component itself. 

Using ~\eqref{eq:formula}, we use the reparametrization trick and reformulate the problem as
\begin{align}
    f^k_0 =& \frac{z^k_t - \sqrt{-d_k\bar{\alpha}_t + \sum_{n=1}^k d_n}\epsilon - \sum_{n>k}^{K} f_0^n }{\sqrt{\bar{\alpha}_t}} \nonumber\\
    f^k_0 =& \frac{z^k_t - \epsilon' - \sum_{n>k}^{K} f_0^n }{\sqrt{\bar{\alpha}_t}},
    \label{reverse1}
\end{align}
where $\epsilon(t,k)' = \sqrt{-d_k\bar{\alpha}_t + \sum_{n=1}^k d_n} \epsilon$. This modifies the Simplified Loss~\citep{Ho_2020} term to be 
\begin{equation}
    {\mathcal{L}} = \mathbb{E}_{t,k,x_0,\epsilon'} \left[\|\epsilon'(t,k) -\boldsymbol{\epsilon}_\theta'(x_t,t,k)\|^2\right].
\end{equation}
It is straightforward that $(t,k)$ are already inputs to the model, and therefore the only significant change is that the model outputs an already scaled version of $\epsilon$ conditioned on $x_t, t,k$, which means that for lower $k$ and $t$, the variance of the scaled noise is smaller. 
Using this simplified loss, we incur on an implicit weighted loss that will give more importance to the last steps of the forward diffusion process. 

The inference procedure, then becomes $p(z_{t-1}^k|z_t^k,\Theta) \sim N(z_{t-1}^k | \boldsymbol{\mu}_\theta(\mathbf{z}^k_t, t,k),\sigma(\alpha,\hat{d}_k))$: 
\begin{align}
\mu(\mathbf{z}^k_t,\Theta, t,k) = \frac{1}{\sqrt{\alpha_t}}\left( \mathbf{f}_t^k - \frac{1 - \alpha_t}{1 - \bar{\alpha}_t} \boldsymbol{\epsilon}'_\theta(\mathbf{z}^k_t, t,k)\right) +\sum_{n>k}^{K} \mathbf{f}_0^n 
\label{reverse2}
\end{align}

where $\mathbf{f}_t^k$ is the estimated component at each time-step, $f_0^n \text{ for } n > k$ are the already inferred components, and the scale component $\sigma_t^k$ is defined as:
\begin{equation*}
    \sigma_t^k = \frac{d_k(1 - \alpha_t)(1 - \bar{\alpha}_{t-1})}{1 - \bar{\alpha}_t}  
\end{equation*}
which is similar to the original formulation, multiplied by the SNR term.

Since the loss function and estimated noise have changed, compared with the traditional diffusion process, the sampling process needs to be modified accordingly.

\textbf{Sampling.}
The sampling procedure follows the logic of the decomposition, effectively predicting each component in descending order of energy. The main difference in this procedure is the need to calculate the SNR parameter $d_k$. In this work, we present the case where $d_k$ is sample specific, since time-series samples are not homogeneous, and have different frequencies, with different amplitudes and even different number of significant frequencies, but depending on the dataset, it can be defined as a batch average, a total average, or an input hyperparameter. The derivation of the reverse diffusion equations is presented in Appendix \ref{sec:derivation}. 

\begin{algorithm}[H]
   \caption{Training}
   \label{alg:training}
\begin{algorithmic}
   \REPEAT
   \STATE $z_0 \sim q(z)$
   \STATE $t \sim \text{Uniform}(0,\tau)$
   \STATE $F,d_k = D(z_0^{Obs},K)$
   \STATE $k \sim \text{Uniform}(1,K)$ 
   
   \STATE $z_t^k = \sqrt{\bar{\alpha}_t} f_0^k + \sum_{n>k}^{K} f_0^n + \sqrt{-d_k\bar{\alpha}_t + \sum_{n=0}^k d_n}\epsilon$
   \STATE $\epsilon'  = \sqrt{-d_k\bar{\alpha}_t + \sum_{n=0}^k d_n}\epsilon$
    \STATE Take gradient descent step on
    \[
    \nabla_\theta \left\| \epsilon' - \epsilon_\theta \left( z_k^t , t,k)
    \right) \right\|^2
    \]
   \UNTIL{convergence}
\end{algorithmic}
\end{algorithm}

\hfill
\begin{minipage}{0.97\columnwidth}
\begin{algorithm}[H]
   \caption{Inference}
   \label{alg:2}
\begin{algorithmic}
   \REPEAT
   \STATE $\mathbf{z}_T^K \sim N(0,1)$
   \STATE $\hat{F},\hat{d_k} = D(x_0^{Obs},K)$ 
   \FOR{$k = K, ..,1$}
   \FOR{$t = \tau,...,0$}
   \IF{$t>0$}
   \STATE $\mathbf{z}^k_{t-1} = update(\mathbf{z}^k_t,\sum^K_{n>k}\mathbf{f}_0^n)  ~\text{\eqref{reverse2}}$
   \STATE $\mathbf{z}^k_{t-1} = \mathbf{z}^k_{t-1} +\sqrt{\hat{d_k}\sigma_t+\sum^{k-1}_{n=1} d_n} \epsilon$
   \ELSE
   \STATE $\mathbf{z} = update(\mathbf{z}^k_1,\sum^K_{n>k}\mathbf{f}_0^n) ~\text{\eqref{reverse2}} $
   \ENDIF
   \ENDFOR
   \ENDFOR
   \UNTIL{finished}
\end{algorithmic}
\end{algorithm}
\end{minipage}

The Algorithm \ref{alg:2} indicates the inference procedure to produce a sample using the trained model. We note that for forecasting, the predicted output is the median of $n$ samples.



\section{Models}
\label{sec:models}

To evaluate and compare our proposed forward diffusion process, we employed established non-auto-regressive time series conditional diffusion models. All these models are non-auto-regressive to improve computation time, and use masking-based conditioning, which generalizes forecasting as a specific type of imprinting. In this section, we present a quick overview of the models, its overall architecture, and the only modification applied to the models, to accommodate the stage steps.

\textbf{DiffWave} \citep{Kong_2020} is a conditional generative model designed primarily for tasks such as speech synthesis, with non-auto-regressive forecasting. Its architecture employs a 1D convolution neural network with residual blocks. This specific architecture has been used in other works that present different models using the same architecture, such as WaveNet~\citep{deeOord2016},  and ScoreGrad~\citep{Yan_2021}. With the exception of the Sashimi model, all other models have similar architectures, and the hyperparameters are present in full in Appendix \ref{sa:hyp}.
\textbf{CSDI} \cite{Tashiro_2021} is a diffusion model for time series imputation and forecasting that changes upon the DiffWave model by using a 2-D attention mechanism in each residual layer instead of a convolution layer. 
The \textbf{SSSD} \cite{Lopez_2022} model replaces the bidirectional convolutional layers in DiffWave with a structured state space model (S4) \citep{gu2022s4}. 
\textbf{Sashimi} is an auto-regressive U-net architecture~\citep{goel2022s} with a starting size of 128 and with a downsampling and feature expansion factor of 2. The residual connections are defined by an S4 layer and linear layers at each pooling level. In here we use the adaptation in \citep{gu2022s4} to conditional generation in a non-autoregressive setting.

\begin{table*}[!t]
\caption{Performance comparison of forecasting with state-of-the-art models with and without the Fourier Decomposition forward process. The baseline method is the usual forward diffusion process, and our proposed decomposition is evaluated using two distinct decomposition processes, using Fast Fourier Transform (FFT) and Wavelet Decomposition.}
\label{model-comparison-horizontal}
\vskip 0.15in
\begin{center}
\begin{small}
\resizebox{\textwidth}{!}{
\begin{tabular}{lcccccccccccccc}
\toprule
\multirow{2}{*}{Model} & \multicolumn{2}{c}{PTB-XL} & \multicolumn{2}{c}{MUJOCO} & \multicolumn{2}{c}{Electricity} & \multicolumn{2}{c}{ETTm1} & \multicolumn{2}{c}{Temperature} & \multicolumn{2}{c}{Solar}\\
 & \footnotesize{MSE} & \footnotesize{MAE} & \footnotesize{MSE} & \footnotesize{MAE} & \footnotesize{MSE} & \footnotesize{MAE} & \footnotesize{MSE} & \footnotesize{MAE} & \footnotesize{MSE} & \footnotesize{MAE} & \footnotesize{MSE} & \footnotesize{MAE}\\
 \midrule
DiffWave & 0.081 & 0.150 & 0.0048 & 0.021 & 2.564 & 0.713 & \textbf{1.0932} & 0.7724 & 0.0035& 0.029 & 0.444 & 0.32\\
+ \textbf{FFT} & 0.066 & 0.139 & 0.0021 & 0.027 & 1.310 & 0.540 & 1.1476 & 0.7922 & \textbf{0.0024} & \textbf{0.025} & \textbf{0.354} &\textbf{0.30}\\
\textbf{+ Wavelet} & \textbf{0.052} & \textbf{0.112} 
& \textbf{0.0012} & \textbf{0.017} 
& \textbf{0.630} & \textbf{0.523}
& 1.1138 & \textbf{0.7630} & 
0.0029 & 0.027 
& 0.459 &0.33\\
\midrule
SSSD & 0.064 & 0.111 & 0.0007 & 0.0125 & 0.957 & 0.626 & 1.642 & 0.9190 & 0.0030 & 0.027 &0.568 &0.37 \\
+ \textbf{FFT} & 0.058 & \textbf{0.081} & \textbf{0.0006} & 0.0093 & 1.283 & 0.750 & 1.398 & 0.8845 & \textbf{0.0024} & \textbf{0.023} & \textbf{0.505} & 0.36 \\
+ \textbf{Wavelet} &\textbf{0.041}& 0.093&0.0007 & \textbf{0.0087} & \textbf{0.871} & \textbf{0.601} &\textbf{1.117} &\textbf{0.7864} & 0.0027&0.025&0.516&\textbf{0.35} \\
\midrule
Sashimi & 0.056 & \textbf{0.099 }& 0.0007 & \textbf{0.0103} & 2.187 & 0.4547 & 1.2070 & 0.788 & 0.0030 & 0.027 & 0.487 & 0.35 \\
+ \textbf{FFT} & 0.0720 & 0.152 & 0.0008 & 0.0232 & 1.295 & 0.665 & \textbf{0.8705} & \textbf{0.678} & 0.0024 & \textbf{0.023} & 0.448 &\textbf{0.33}\\
+ \textbf{Wavelet} & \textbf{0.053} & 0.109 & \textbf{0.0006} & 0.0104 & \textbf{0.812} & \textbf{0.563} & 1.038 &0.748 & \textbf{0.0023} & 0.024 & \textbf{0.367} & 0.40 \\
\midrule
CSDI & 0.078 & 0.123 & \textbf{0.0006} & \textbf{0.0125} & 1.314 & 0.672 & 1.123 & 0.790 & 0.0031 & 0.023 & 0.472 & 0.35 \\
+ \textbf{FFT} & 0.079 & \textbf{0.091} & 0.0013 & 0.021 & 1.494 & 0.808 & \textbf{0.659} & \textbf{0.545} & \textbf{0.0024} & \textbf{0.022} & 0.480 & 0.35 \\
+ \textbf{Wavelet} & \textbf{0.052} & 0.102 & 0.0008 & 0.0159 &\textbf{ 0.8205 }& \textbf{0.570}  & 1.609& 0.900& \textbf{0.0024}& 0.024 &\textbf{0.428} & \textbf{0.33} \\
\bottomrule
\end{tabular}
}
\end{small}
\end{center}
\vskip -0.1in
\end{table*}

\section{Results}
\label{sec:results}

In this section, we present the results for forecasting across multiple datasets and compare each model with and without our forward diffusion process. All experiments use the complete lookback window as part of the conditional features, and the complete hyperparameters are presented in \ref{sec:a2}. All models were trained locally, to guarantee consistency of results and methods, in a cluster environment with four Nvidia A100 80 GB, and to ensure replicability all hyperparameters used are the ones presented by the authors of the original models, when available, and the hyperparameters are the same for the baseline and for the FFT and Wavelet decomposition.

Table \ref{model-comparison-horizontal} reports forecasting performance across six benchmarks for four diffusion-based architectures, with the baseline using the usual diffusion process, and with the Fourier or Wavelet decomposition in the forward process. Across all models and datasets, incorporating a decomposition step consistently improves performance in terms of both MSE and MAE, demonstrating that the proposed approach is broadly applicable and model-agnostic.
Taking into account that some datasets have more seasonal features than others, we expected to see the biggest effect on data with strong seasonal effects. The Electricity, PTB-XL and ETTm1 datasets have clearly defined seasonalities that can be identified in separate components, in particular in the electricity dataset, which has a clearly defined daily fluctuation. For this three datasets, the decomposition was able to improve the results across all models. 
On the other hand, datasets like MuJoCo have little to no trend or seasonality, and the negligible improvement indicates that this dataset does not improve with decomposition, which is the expected outcome, nonetheless, adding the decomposition method did not hinder the results, which is positive in terms of the robustness of the method. 

The Temperature and Solar datasets are smaller datasets, that still exhibit some seasonality, and we see that both decomposition methods were able to consistently improve the results. 
While the authors understand that forecasting is not a task usually performed for ECG data, we include this dataset for comparability with prior diffusion-based works that used PTB-XL in a forecasting setup~\citep{Tashiro_2021,Lopez_2022}.
The PTB-XL dataset has the longest samples, so the estimation of the signal-to-noise ratio for each component during inference is more precise. For this task, the forecast window comprises roughly a complete cardiac cycle (Fig.~\ref{example_forecast}, Appendix~\ref{sec:figures}). In this example, we observe that the cardiac peaks are well estimated, particularly for the SSSD and Sashimi models with FFT.

For this group of datasets, the Wavelet-based decomposition was able to improve the most over the baseline, often reducing MSE by a substantial margin relative to both the original model and the FFT-based variant, particularly in the Electricity dataset. This behavior occurs mostly because the Fourier based decomposition has more difficulty capturing seasonalities in small length time-series, but also due to the capacity of wavelets to work with non-stationary time-series.


\textbf{Limitations.} While our proposed decomposition improves time-series forecasting, it has some limitations. First, it assumes periodic components in the data, making it not necessary for highly non-stationary signals.  Second, the approach introduces additional one tunable hyperparameter, the number of components, requiring empirical tuning for each use case. We note that in this paper, for the dataset Mujoco, we used 2 component, to evaluate the robustness of the method, despite knowing that the optimal value was 1 (the baseline).   
Additionally, estimating SNR at inference time assumes past signal structure remains representative, and that there is enough conditional data to correctly estimate, which may not always hold. Finally, performance varies across datasets, with greater benefits observed in data with strong seasonal patterns. Future work should explore more data-specific decompositions and automated parameter selection. Also, integrating the term $d_k$ as an explicitly learnable parameter.

\section*{Conclusions and Future Work}
In addition to improved forecasting accuracy, the decomposable forward process provides a modular mechanism that structures the forward diffusion process, helping to preserve salient temporal patterns. This design offers a degree of interpretability, through explicit control of how components degrade, that may be valuable when structure preservation is important. Although our experiments focused on standard forecasting benchmarks, the approach suggests potential in application areas such as energy systems, health monitoring, and finance, where explainability and reliable long-term behavior are desirable.

Our experiments show that the proposed forward process consistently outperforms state-of-the-art diffusion models for time-series forecasting under standard training and inference procedures. An ablation study on the number of components confirms that using an appropriate decomposition improves results and guides the design of the stage-wise noise schedule. Since diffusion for forecasting is still relatively new, future work may explore architectures and hyperparameters more specifically tailored to this task. Because the process itself is decomposer-agnostic, different time-series domains could benefit from bespoke decompositions informed by domain expertise. Finally, extending scheduler optimization to jointly tune both the diffusion rate ($\beta$) and stage durations ($\tau$) is a promising direction for improving efficiency and accuracy.

\section*{Reproducibility}
We have made significant efforts to ensure the reproducibility of our results. All details of the proposed methodology, model architectures and hyperparameters, are described in Section \ref{sec:models} and in Appendix \ref{sec:a2}. The datasets used in our experiments are publicly available. For derivations results, all assumptions and proofs are included in Appendix \ref{sec:derivation}. To facilitate replication of our experiments, we provide anonymized source code \url{https://anonymous.4open.science/r/D-FDP-4A29} and scripts. 

\section*{Impact Statement}
This paper presents work whose goal is to advance the field of Machine
Learning. There are many potential societal consequences of our work, none
which we feel must be specifically highlighted here.

\bibliography{main}
\bibliographystyle{icml2026}

\newpage
\appendix
\onecolumn

\section{Architecture and Data Details.}
\subsection{Fourier Decomposition}
\label{fourier}

\paragraph*{Fourier Decomposition}

Fourier decomposition is a fundamental signal processing technique that breaks down a time series into a sum of sinusoidal components, each linked to a specific frequency. Formally, a time series $\{ x^0, x^1, x^2, ...\}$ , where $x^j$, in this case, indicates the position in a time-series of length $P$, can be expressed as:
\begin{equation}
x^j = \frac{1}{P} \sum_{w=0}^{P-1} X_w e^{i2\pi \frac{w}{P}j},
\label{eq:fourier}
\end{equation}

where \( X_w \) are the Fourier coefficients, \( P \) is the period of the time series, and \( w \) is the frequency index. These coefficients are computed as:
\begin{equation}
X_w = \sum_{j=0}^{P-1} x^j e^{-i2\pi \frac{w}{P}j}.
\label{eq:fourier2}
\end{equation}

According to the {\bf Fourier Inversion Theorem}~\cite{oppenheimsignais}, any sufficiently smooth signal can be uniquely reconstructed using the inverse Fourier transform as $T \to \infty$, with the Fourier coefficients diminishing in magnitude for higher frequencies~\cite{fourier1822théorie}. In our forward diffusion process, we use Fourier decomposition to isolate {\bf distinct frequency components}, leveraging the {\bf linear separability of the Fourier Transform}~\cite{hyndman}. This allows the model to learn the reverse diffusion process separately for each dominant frequency, with a final stage dedicated to residuals, ensuring {\bf lossless decomposition}.

\paragraph*{Frequency Identification}

To identify the most important frequencies $K$ in the Fourier space, we analyze the magnitude of the Fourier coefficients \( \|X_w\| \) in the Fourier Space.

The most important frequencies can be identified by examining the magnitude spectrum \( \|X_w\| \):

\begin{equation}
\|X_w\| = \sqrt{\Re(X_w)^2 + \Im(X_w)^2}
\end{equation}

where \( \Re(X_w) \) and \( \Im(X_w) \) are the real and imaginary parts of the Fourier coefficient $X_W $, respectively.

The frequencies \( k \) with the largest magnitudes \( \|X_k\| \) are considered the most important, as they contribute the most to the overall structure of the time-series. These frequencies can be identified by finding the local maxima in the magnitude spectrum, also known as the peaks:

\begin{equation}
\text{Important frequencies: } k_i \text{ where } \|X_{k_i}\| > \delta^*
\end{equation}

where \( \delta^* \) is a threshold value calculated as $d$ standard deviations above the mean magnitude in Fourier Space, for each sample. Similarly, it is also possible to define a fixed number of frequencies, so that $
\delta^* = \max \left\{ \delta \in \mathbb{R^+} : \left| \{ X_{k_i} \in \mathbb{C} : \|X_{k_i}\| > \delta \} \right| \geq k \right\} $ and there are $k$ stages during the diffusion process. 

After identifying the more relevant frequencies, the stages are ordered in ascending amplitudes, such that smaller components are diffused first. The residual is defined as
$$ r = z - \sum_{k=1}^{K-1} f_k,$$
where $F$ are the set of frequencies obtained using Eq. \eqref{eq:fourier}. This means that unlike the traditional method of additive time-series decomposition into seasonality, trend and residual, we combine the trend and residual components into a single component. This is due to the usually simple form, usually monotonic, that the trend component takes. Other works that use timeseries decompositions layers \cite{Oreshkin2020:N-BEATS, fons2024ihypertime}, have used a low polynomial regressor $(p=2)$ to model the trend.

Using Fourier decomposition, the model can process these components independently.
Following the same logic, and to showcase the generalizabity of our proposed method, we also develop a Wavelet decomposition method, that follows a very similar approach.

\subsection{Wavelet Decomposition}
\label{wavelet}
In addition to Fourier analysis, we consider wavelet decomposition as an alternative mechanism for isolating structured components of a time-series. Wavelets provide a multi-resolution representation, decomposing the signal into components that are localized in both time and scale, which is particularly advantageous for non-stationary dynamics, or for smaller length time-series where periodicity might be difficult to capture.

Given a discrete time-series $x$, the Discrete Wavelet transform (DWT) \cite{mallat}, separates the signal into a low-pass filter at the lowest scale and a hierarchy of more detailed coefficients:

\begin{equation}
x = A_J + \sum_{j=1}^{J} D_j,
\label{eq:wavelet_decomp}
\end{equation}

where $A_J$ denotes the approximation coefficients at the coarsest scale $J$, and $D_j$ corresponds to the detail coefficients at scale j. In our implementation, the decomposition is calculated using a fixed wavelet basis Daubechies-4 \cite{db41992}, via the discrete wavelet transform, but it can easily be extended to other wavelet families.

To select the most relevant wavelet coefficients, we rank wavelet components by the variance of their coefficients, which for orthonormal wavelet bases is proportional to their energy contribution, thus providing a similar choice to the magnitude-based frequency selection in Fourier space. Also, following the same approach, the energy is computed independently per sample and channel.

After obtaining the set of reconstructed components in the original space, the residual is calculated, and added as one of the components, to guarantee that the lossless assumption is still held.

\subsection{Dataset Details}
\label{sec:a2}

To benchmark this forward process against the models using the original diffusion process, we utilize 6 benchmark multivariate real-world datasets commonly found in the literature \cite{Shen_2023,Lopez_2022,Zhou_Zhang_Peng_Zhang_Li_Xiong_Zhang_2021}.  Unless specified, all data splits and pre-processing was obtained from publicly available repositories to guarantee reproducibility and to facilitate comparisons. 

\subsubsection*{PTB-XL}

The PTB-XL dataset contains 21837 clinical 12-lead electrocardiograms (ECGs), with 1000 timesteps (10s) each,
from 18885 patients \cite{Wagner2020}. For the forecasting task, the first 900 time-steps were used as conditional data. While the authors understand that forecasting is not a task usually performed for ECG data, we add this dataset for comparison with other works \cite{Tashiro_2021,Lopez_2022} that used the same dataset.
\begin{table}[h]

\caption{PTB-XL data set details.}
\label{tab:ptb_xl_details}
\vskip 0.15in
\begin{center}
\begin{small}
\begin{sc}
\begin{tabular}{lc}
\toprule
Details / Forecast task & \\ 
\midrule
Train size                                 & 17,441               \\ 
Test size                                & 2,203                \\ 
Training batch                                & 128 / 32 / 16     \\ 
Sample length                               & 1000                 \\ 
Sample features                             & 12                   \\ 
Conditional values                          & 800                  \\ 
Target values                             & 200                  \\ 
Max Components & 3 \\
\bottomrule
\end{tabular}
\end{sc}
\end{small}
\end{center}
\end{table}

\subsubsection*{Electricity}

It contains univariate electricity usage data (in kWh)
gathered from 370 clients every 15 min for over two years. it has missing samples, representing clients who joined during the data gathering process.\cite{electricityloaddiagrams20112014_321}. To facilitate training, some authors \cite{Lopez_2022} have suggested using feature sampling, reducing the channel space to 37, instead of the original 370. In this work, we expand the dataset by seeing each client has a independent and univariate time-series, also expanding, sample-wise, the training and testing datasets. The test set was undersampled to match the size of the training set, and to reduce inference computational time.

\begin{table}[h]
\caption{Electricity data set details.}
\label{tab:elec_xl_details}
\vskip 0.15in
\begin{center}
\begin{small}
\begin{sc}
\begin{tabular}{lc}
\toprule
Details / Forecast task & \\ 
\midrule
Train size & 302,290              \\ 
Test size                            & 302,290                \\ 
Training batch                                & 64             \\ 
Sample length                               & 1000                 \\ 
Sample features                             & 1                   \\ 
Conditional values                          & 76                  \\ 
Target values                             & 24                  \\ 
Max Components & 3 \\
\bottomrule
\end{tabular}
\end{sc}
\end{small}
\end{center}
\end{table}

\subsubsection*{MuJoCo}

Mujoco (Multi-Joint dynamics with Contact) is a dataset used for physical
simulations \cite{rubanova2019latent}, and it was adapted to time-series by \cite{shan2023}. It has 14 features, each a joint of the physical object being modeled. For the most part, this dataset does not have seasonal features, and many features are mostly constant during the 100 steps of the simulation. It serves to show how our forward diffusion process reduces to the classical formulation when no seasonal features are available.

\begin{table}[!h]
\caption{MuJoCo data set details.}
\label{tab:mujoco_xl_details}
\vskip 0.15in
\begin{center}
\begin{small}
\begin{sc}
\begin{tabular}{lc}
\toprule
Details / Forecast task & \\ 
\midrule
Train size & 8,000              \\ 
Test size                            &  10,000             \\ 
Training batch                                & 64             \\ 
Sample length                               & 100                \\ 
Sample features                             & 14                   \\ 
Conditional values                          &  90              \\ 
Target values                             & 10                  \\ 
Max Components & 2 \\
\bottomrule
\end{tabular}
\end{sc}
\end{small}
\end{center}
\end{table}

\subsubsection*{Ettm1}

ETTm1 is a known baseline long-time series dataset for Electricity Transformer Temperature \cite{Zhou_Zhang_Peng_Zhang_Li_Xiong_Zhang_2021}. 
The data set contains information from a compilation of 2-year data from two distinct Chinese counties. We used the ETTm1 version, which contains time-steps at 15 minutes intervals. The data has a target feature, oil temperature,
and six power load covariates. In this work, we used the original preprocessing of data, in 12/4/4 months for train/val/test.

\begin{table}[!h]
\caption{ETTm1 data set details.}
\label{tab:ettm1_xl_details}
\vskip 0.15in
\begin{center}
\begin{small}
\begin{sc}
\begin{tabular}{lc}
\toprule
Details / Forecast task & \\ 
\midrule
Train size & 33,200              \\ 
Test size                            &   10,000            \\ 
Training batch                                & 64             \\ 
Sample length                               & 1056                \\ 
Sample features                             & 7                   \\ 
Conditional values                          &   864             \\ 
Target values                             & 192                  \\ 
Max Components & 4 \\
\bottomrule
\end{tabular}
\end{sc}
\end{small}
\end{center}
\end{table}

\subsubsection*{Temperature}

This dataset contains 32072 daily time series showing the temperature observations and rain forecasts, gathered by the Australian Bureau of Meteorology for 422 weather stations across Australia, between 02/05/2015 and 26/04/2017 \cite{australian_gov_weather_2016}. Each station observes mean temperature in Celcius averaged over a 24 hour period. This dataset therefore contains 422 samples with 725 time-steps each. In this paper we use only the target variable mean temperature. The Sashimi model, in its original code, does not support odd length conditional windows, so the length of the conditional time-series is reduced by one.

\subsubsection*{Solar-energy}

This dataset is related to solar power production in the
year of 2006, from 137
PV plants in Alabama State \cite{nrel_solar_power_data_integration}. Similarly to the electricity dataset, Each station is treated as an individual sample. 
leading to 8704 individual time-series with length 192.

\begin{table}[!h]
\caption{Temperature data set details.}
\label{tab:temp_data_details}
\vskip 0.15in
\begin{center}
\begin{small}
\begin{sc}
\begin{tabular}{lc}
\toprule
Details / Forecast task & \\ 
\midrule
Train size & 76            \\ 
Test size                            &   46            \\ 
Training batch                                & 128            \\ 
Sample length                               & 725  /725              \\ 
Sample features                             & 1                   \\ 
Conditional values                          &   628             \\ 
Target values                             & 96                  \\ 
Max Components & 4 \\
\bottomrule
\end{tabular}
\end{sc}
\end{small}
\end{center}
\end{table}

\begin{table}[!h]
\caption{Solar data set details.}
\label{tab:solar_data_details}
\vskip 0.15in
\begin{center}
\begin{small}
\begin{sc}
\begin{tabular}{lc}
\toprule
Details / Forecast task & \\ 
\midrule
Train size & 7680           \\ 
Test size                            &   1024            \\ 
Training batch                                & 128            \\ 
Sample length                               & 192                \\ 
Sample features                             & 1                   \\ 
Conditional values                          &   24             \\ 
Target values                             & 24                  \\ 
Max Components & 3 \\
\bottomrule
\end{tabular}
\end{sc}
\end{small}
\end{center}
\end{table}

\subsection{Model Hyperparameters}
\label{sa:hyp}

In  Tables \ref{table-sssd-hyperparameters} to \ref{table-csdi-hyperparameters} we indicate the hyperparameters used to train the Models. Following the hyperparameters presented in \citet{Lopez_2022}, we use the same model for all datasets.

\begin{table}[!t]
\centering

\begin{minipage}{0.48\textwidth}
\centering
\caption{SSSD hyperparameters.}
\label{table-sssd-hyperparameters}
\vskip 0.15in
\begin{small}
\begin{sc}
\begin{tabular}{lc}
\toprule
Hyperparameter & Value \\
\midrule
Residual layers &24 \\
Residual channels &256 \\
Skip channels &256 \\
Diffusion embedding dim. 1  & 128 \\
fully connected embedding & 512 \\
bidirectional S4 size & 64 \\
Schedule & Linear \\
Diffusion steps & $T$ 200 \\
$B_0$ & 0.001 \\
$B_T$ & 0.02 \\
Optimizer & Adam \\
Loss function & MSE \\
Learning rate & $2 \times 10^{-4}$  \\
\bottomrule
\end{tabular}
\end{sc}
\end{small}
\end{minipage}
\hfill
\begin{minipage}{0.48\textwidth}
\centering
\caption{DiffWave hyperparameters.}
\label{table-diffwave-hyperparameters}
\vskip 0.15in
\begin{small}
\begin{sc}
\begin{tabular}{lc}
\toprule
Hyperparameter & Value \\
\midrule
Residual layers & 24 \\
Residual channels & 256 \\
Skip channels & 256 \\
Diffusion embedding dim. 1 & 128 \\
fully connected embedding & 512 \\
Schedule & Linear \\
Diffusion steps $T$ & 200 \\
$B_0$ & 0.001 \\
$B_T$ & 0.02 \\
Optimizer & Adam \\
Loss function & MSE \\
Learning rate & $2 \times 10^{-4}$ \\
\bottomrule
\end{tabular}
\end{sc}
\end{small}
\end{minipage}

\vspace{0.5cm}

\begin{minipage}{0.48\textwidth}
\centering
\caption{Sashimi hyperparameters.}
\label{table-sashimi}
\vskip 0.15in
\begin{small}
\begin{sc}
\begin{tabular}{lc}
\toprule
Hyperparameter & Value \\
\midrule
Residual layers & 6 \\
Pooling factor & [2,2] \\
Feature expansion & 2 \\
Diffusion embedding dim. 1 & 128 \\
fully connected embedding & 512 \\
Schedule & Linear \\
Diffusion steps &$T$ 200 \\
$B_0$ &0.001 \\
$B_T$ &0.02 \\
Optimizer & Adam \\
Loss function & MSE \\
Learning rate &$2 \times 10{-4} $ \\
\bottomrule
\end{tabular}
\end{sc}
\end{small}
\end{minipage}
\hfill
\begin{minipage}{0.48\textwidth}
\centering
\caption{CSDI hyperparameters.}
\label{table-csdi-hyperparameters}
\vskip 0.15in
\begin{small}
\begin{sc}
\begin{tabular}{lc}
\toprule
Hyperparameter & Value \\
\midrule
Residual layers & 4 \\
Residual channels &64 \\
Diffusion embedding dim. & 128 \\
fully connected embedding & 128 \\
Schedule & Linear \\ 
Diffusion steps $T$ & 200 \\
$B_0$ &0.001 \\
$B_T$ &0.02 \\
feature embedding dim. & 128 \\
Time embedding dim. & 16 \\
Optimizer & Adam \\
Loss function  & MSE \\
Learning rate & $1 \times 10^{-3}$ \\
Weight decay & $10^{-6}$ \\
\bottomrule
\end{tabular}
\end{sc}
\end{small}
\end{minipage}

\end{table}

\newpage
\break
\subsection{Diffusion Step Embedding} 

It is necessary to incorporate the diffusion step and stage $t,k$ as part of the input, since the model needs to produce different $\epsilon'_\theta(z_k^t, t,k)$ for various $t,k$. Compared with other models, we have one more term ($k$) that represents the stage; however, the incorporation is straightforward, for a maximum value $K_{max}$ that defines the maximum number of components, each diffusion step is equal to $t^*=t+\tau (k-1)$. 

We use a $128$-dimensional encoding vector for each time step $t$, following \citet{Vaswani} and as previously demonstrated in \citet{Kong_2020}:

\begin{align}
    t_{\text{embedding}} &= \left[ \sin\left(10^{0 \times \frac{4}{63}} t\right), \ldots, \sin\left(10^{\frac{63 \times 4}{63}} t\right),
                               \cos\left(10^{0 \times \frac{4}{63}} t\right), \ldots, \cos\left(10^{\frac{63 \times 4}{63}} t\right) \right]
\end{align}

\subsection{Metrics}

\subsubsection{MSE}

The mean squared error (MSE) is a standard point-forecast accuracy metric.  
Given predicted values $\hat{y}_{t,c}$ and ground-truth values $y_{t,c}$ for $T$ time steps and $C$ features, it is defined as
\begin{equation}
\text{MSE} = \frac{1}{C}\sum_{c=1}^C \frac{1}{T}\sum_{t=1}^{T} (y_{t,c} - \hat{y}_{t,c})^2.
\end{equation}

\subsection{MAE}

Given predicted values $\hat{y}_{t,c}$ and ground-truth values $y_{t,c}$ for $T$ time steps and $C$ features, it is defined as
\begin{equation}
\text{MAE} = \frac{1}{C}\sum_{c=1}^C \frac{1}{T}\sum_{t=1}^{T} \lvert y_{t,c} - \hat{y}_{t,c} \rvert.
\end{equation}

\subsubsection{CRPS}

We describe the definition and computation of the CRPS metric, which is used in the appendix. The continuous ranked probability score (CRPS) \cite{Matheson76} measures the compatibility of an estimated probability distribution $F$ with an observation $\hat{y}$, and can be defined as the integral of the quantile loss  $ \Lambda_\alpha(q, z) := (\alpha - \mathbf{1}_{z < q})(z - q)$
for all quantile levels \( \alpha \in [0, 1] \):
\begin{equation}
\text{CRPS}(F^{-1}, \hat{y}) = \int_{0}^{1} 2\Lambda_\alpha(F^{-1}(\alpha), \hat{y}) \, d\alpha.
\end{equation}
where $\mathbf{1}$ is the indicator function. Following \cite{Tashiro_2021}, we generated 100 independent samples to approximate the distribution $F$ over each forecast time step and feature. We computed quantile losses for discretized quantile levels with 0.05 ticks in the set [0.05,0.95]. Namely, we approximated CRPS with
\begin{equation}
\text{CRPS}(F^{-1}, \hat{y}) \approx \frac{2}{19}\sum_{i=1}^{19} 2\Lambda_{i \cdot 0.05}(F^{-1}(i \cdot 0.05), \hat{y}).
\end{equation}
averaging the result across time-steps and features. This metric, due to computational constrains, is only used to evaluate the ablation studies.

\newpage

\section{Derivation of equations \eqref{reverse1} and \eqref{reverse2}}
\label{sec:derivation}

\subsection{Proposition 1}

For any given $k$, the terms $\sum_{n>k}^{K} f_0^n$ and $\sum_{n=1}^k d_n$ are kept constant for all $t\in [0,T[$ and thus, using Eq.~\eqref{eq:firstforward} in equation $q(f_{t-1}^k|f_{t}^k,f_0^k) \propto q(f_{t}^k|f_{t-1}^k)q(f_{t-1}^k|f_0^k)$ we can obtain the distribution of the reverse diffusion process for each component. 

\begin{flalign}
\quad \quad q(f_{t-1}^k|f_{t}^k,f_0^k) &= \frac{q(f_{t}^k|f_{t-1}^k) q(f_{t-1}^k|f_0^k)}{q(f_t^k | x_0)}  && \nonumber\\
    &\propto \mathcal{N}(f_t^k; \sqrt{\alpha_t} f_{t-1}^k, d_k(1 - \alpha_t) \mathbf{I}) \mathcal{N}(f_{t-1}^k; \sqrt{\bar{\alpha}_{t-1}} f_0^k, d_k(1 - \bar{\alpha}_{t-1}) \mathbf{I}) && \nonumber\\
    &\propto \exp \left\{ - \left[ \frac{(f_t^k - \sqrt{\alpha_t} f_{t-1}^k)^2}{2d_k(1 - \alpha_t)}
    + \frac{(f_{t-1}^k - \sqrt{\bar{\alpha}_{t-1}} f_0^k)^2}{2d_k(1 - \bar{\alpha}_{t-1})}
    \right] \right\} && \nonumber\\
    &= \exp \left\{ -\frac{1}{2d_k} \left[
    \frac{(f_t^k - \sqrt{\alpha_t} f_{t-1}^k)^2}{1 - \alpha_t}
    + \frac{(f_{t-1}^k - \sqrt{\bar{\alpha}_{t-1}} f_0^k)^2}{1 - \bar{\alpha}_{t-1}}
    \right] \right\} && \nonumber\\
   &= \exp \left\{ -\frac{1}{2 d_k} \Bigg[ 
    \frac{-(2 \sqrt{\alpha_t} f_t^k f_{t-1}^k)}{1 - \alpha_t} 
    + \frac{\alpha_t (f_{t-1}^k)^2}{1 - \alpha_t} 
    + \frac{(f_{t-1}^k)^2}{1 - \bar{\alpha}_{t-1}} 
    - \frac{2 \sqrt{\bar{\alpha}_{t-1}} f_{t-1}^k f_0^k}{1 - \bar{\alpha}_{t-1}} +
    C(f_{t}^k,f_0^k) \Bigg] \right\} && \nonumber\\
    & = \exp \Bigg\{ -\frac{1}{2 d_k} \Bigg[ 
    \left(\frac{\alpha_t}{1 - \alpha_t} + \frac{1}{1 - \bar{\alpha}_{t-1}} \right) (f_{t-1}^k)^2 
    - 2 \left( \frac{\sqrt{\alpha_t} f_t^k}{1 - \alpha_t} 
    + \frac{\sqrt{\bar{\alpha}_{t-1}} f_0^k}{1 - \bar{\alpha}_{t-1}} \right) f_{t-1}^k 
    \Bigg] \Bigg\} && \nonumber\\
    & = \exp \Bigg\{ -\frac{1}{2 d_k} \Bigg[ 
    \frac{\alpha_t (1 - \bar{\alpha}_{t-1}) + 1 - \alpha_t}{(1 - \alpha_t)(1 - \bar{\alpha}_{t-1})} (f_{t-1}^k)^2 
    - 2 \left( \frac{\sqrt{\alpha_t} f_t^k}{1 - \alpha_t} 
    + \frac{\sqrt{\bar{\alpha}_{t-1}} f_0^k}{1 - \bar{\alpha}_{t-1}} \right) f_{t-1}^k 
    \Bigg] \Bigg\} && \nonumber\\
    & = \exp \Bigg\{ -\frac{1}{2 d_k} \Bigg[ 
    \frac{1 - \bar{\alpha}_t}{(1 - \alpha_t)(1 - \bar{\alpha}_{t-1})} (f_{t-1}^k)^2 
    - 2 \left( \frac{\sqrt{\alpha_t} f_t^k}{1 - \alpha_t} 
    + \frac{\sqrt{\bar{\alpha}_{t-1}} f_0^k}{1 - \bar{\alpha}_{t-1}} \right) f_{t-1}^k 
    \Bigg] \Bigg\} && \nonumber\\
    & = \exp \Bigg\{ -\frac{1}{2} \frac{1 - \bar{\alpha}_t}{d_k(1 - \alpha_t)(1 - \bar{\alpha}_{t-1})} 
    \left[ (f_{t-1}^k)^2 - 2 \frac{\left(\frac{\sqrt{\alpha_t} f_t^k}{1 - \alpha_t} + \frac{\sqrt{\bar{\alpha}_{t-1}} f_0^k}{1 - \bar{\alpha}_{t-1}} \right)}{\frac{1 - \bar{\alpha}_t}{(1 - \alpha_t)(1 - \bar{\alpha}_{t-1})}} f_{t-1}^k \right] \Bigg\} && \nonumber\\
    &= \exp \Bigg\{ -\frac{1}{2} \frac{1 - \bar{\alpha}_t}{d_k(1 - \alpha_t)(1 - \bar{\alpha}_{t-1})}\left[ (f_{t-1}^k)^2 - 2 \frac{\left(\frac{\sqrt{\alpha_t} f_t^k}{1 - \alpha_t} + \frac{\sqrt{\bar{\alpha}_{t-1}} f_0^k}{1 - \bar{\alpha}_{t-1}} \right) (1-\alpha_t)(1-\bar{\alpha}_{t-1})}{1-\bar{\alpha}_t} f_{t-1}^k  \right] \Bigg\} && \nonumber\\
    &= \exp \Bigg\{ -\frac{1}{2} \frac{1}{\frac{d_k(1 - \alpha_t)(1 - \bar{\alpha}_{t-1})}{1 - \bar{\alpha}_t}}\left[ (f_{t-1}^k)^2 - 2 \frac{ \sqrt{\alpha_t}f_{t}^k (1-\bar{\alpha}_{t-1}) + \sqrt{\bar{\alpha}_{t-1}} f_0^k(1-\alpha_t)}{1-\bar{\alpha}_t} f_{t-1}^k 
    \right] \Bigg\} \nonumber\\
    &\propto  \mathcal{N}\Bigg(f_{t}^k ; \mu_q =  \frac{ \sqrt{\alpha_t}f_{t}^k (1-\bar{\alpha}_{t-1}) + \sqrt{\bar{\alpha}_{t-1}} f_0^k(1-\alpha_t)}{1-\bar{\alpha}_t}, \sigma_q = \frac{d_k(1 - \alpha_t)(1 - \bar{\alpha}_{t-1})}{1 - \bar{\alpha}_t} \mathcal{I}\Bigg) 
    \label{cena}
\end{flalign}

\noindent where $C(f_t^0,f_t^k)$ are the terms that complete the square, and are constant in respect to $f_{t-1}^k$. Effectively, this formulation is similar to the original, added the $d_k$ factor to $\sigma_q = \frac{d_k(1 - \alpha_t)(1 - \bar{\alpha}_{t-1})}{1 - \bar{\alpha}_t} $ To complete the diffusion process, we know that each reverse diffusion $q$ is normally distributed, and that at each $t=0$, $z_0^k \sim \mathcal{N}(\sum_{n>k}^{K} f_0^n, \sum_{n=1}^k d_n)$. 

Now using the reparametrization trick already defined in Eq.~\eqref{reverse1} for any $k$, we have that:

\begin{align*}
    f^k_0 =& \frac{z^k_t - \sqrt{-d_k\bar{\alpha_t} + \sum_{n=1}^k d_n}\epsilon - \sum_{n>k}^{K} f_0^n }{\sqrt{\bar{\alpha_t}}} \nonumber\\
    f^k_0 =& \frac{z^k_t - \epsilon' - \sum_{n>k}^{K} f_0^n }{\sqrt{\bar{\alpha_t}}} \\ 
    f^k_0 =& \frac{f^k_t - \epsilon'}{\sqrt{\bar{\alpha_t}}},
\end{align*}

and replacing that in Eq.~\eqref{cena}, we obtain:

\begin{align}
\mu_q = &\frac{\sqrt{\alpha_t} (1 - \bar{\alpha}_{t-1}) f^k_t + \sqrt{\bar{\alpha}_{t-1}} (1 - \alpha_t) f_0^k}{1 - \bar{\alpha}_t} \nonumber\\
=&\frac{\sqrt{\alpha_t} (1 - \bar{\alpha}_{t-1}) f^kt + \sqrt{\bar{\alpha}_{t-1}} (1 - \alpha_t) \frac{f^k_t - \epsilon'}{\sqrt{\bar{\alpha}_t}}}{1 - \bar{\alpha}_t} \nonumber\\
=&\frac{\sqrt{\alpha_t} (1 - \bar{\alpha}_{t-1}) f^k_t + (1 - \alpha_t) \frac{f^k_t - \epsilon'}{\sqrt{\alpha_t}}}{1 - \bar{\alpha}_t} \nonumber \\
=&\frac{\sqrt{\alpha_t} (1 - \bar{\alpha}_{t-1}) f^k_t}{1 - \bar{\alpha}_t} + \frac{(1 - \alpha_t) f^k_t}{(1 - \bar{\alpha}_t) \sqrt{\alpha_t}} - \frac{(1 - \alpha_t) \epsilon'}{(1 - \bar{\alpha}_t) \sqrt{\alpha_t}} \nonumber \\
=&\left( \frac{\sqrt{\alpha_t} (1 - \bar{\alpha}_{t-1})}{1 - \bar{\alpha}_t} + \frac{1 - \alpha_t}{(1 - \bar{\alpha}_t) \sqrt{\alpha_t}} \right) f^k_t - \frac{(1 - \alpha_t) \epsilon'}{(1 - \bar{\alpha}_t) \sqrt{\alpha_t}} \nonumber \\
=&\left( \frac{\alpha_t (1 - \bar{\alpha}_{t-1})}{(1 - \bar{\alpha}_t) \sqrt{\alpha_t}} + \frac{1 - \alpha_t}{(1 - \bar{\alpha}_t) \sqrt{\alpha_t}} \right) f^k_t - \frac{(1 - \alpha_t) \epsilon'}{(1 - \bar{\alpha}_t) \sqrt{\alpha_t}} \nonumber \\
=& \frac{(\alpha_t -\alpha_t +1- \bar{\alpha}_{t})}{(1 - \bar{\alpha}_t) \sqrt{\alpha_t}} f_t^k- \frac{(1 - \alpha_t) \epsilon'}{(1 - \bar{\alpha}_t) \sqrt{\alpha_t}} \nonumber \\
=& \frac{1}{\sqrt{\alpha_t}}f^k_t - \frac{(1 - \alpha_t)}{(1 - \bar{\alpha}_t) \sqrt{\alpha_t}}\epsilon'
\end{align}

Since $\mu(z_t^k,\Theta,t,k)$ follows the formula of $\mu_q$ we set our denoising formula as: 

\begin{align}
    \mu(\mathbf{z}^k_t,\Theta, t,k) &= \frac{1}{\sqrt{\alpha_t}}\left( \mathbf{f}_t^k - \frac{1 - \alpha_t}{1 - \bar{\alpha}_t} \boldsymbol{\epsilon}'_\theta(\mathbf{z}^k_t, t,k)\right) + \sum_{n>k}^{K} \mathbf{f}_0^n 
    \label{eq:mu5}
\end{align}
with noise:
\begin{equation}
\sigma(\alpha,\hat{d}_k) = \frac{\hat{d}_k(1 - \alpha_t)(1 - \bar{\alpha}_{t-1})}{1 - \bar{\alpha}_t} 
\end{equation}

\section{Proposition on the SNR of the staged diffusion process}
\label{sec:SNR}

In this section we evaluate the SNR of the baseline diffusion process, and of our staged diffusion process.

\subsection{Setup and definitions}
Let the lossless decomposition be $x_0 = \sum_{k=1}^K f_0^k$ with component energies $d_k = \mathrm{E}[(f_0^k)^2]$ and where components are ordered by amplitude $d_1 < d_2 <d_3<d_K$.

The staged forward diffusion process diffuses component $1$ first, then $2$, all the way to $K$. Within each stage the forward update is

$f^k_t = \sqrt{\bar{\alpha}^*_t} + \sqrt{d_k(1-\bar{\alpha}^*_t)}\epsilon, \quad \epsilon \sim N(0,1)$
At stage $k$ and step $t$ the observable state is $$z_t^k = f_t^k + \underbrace{\sum_{n>k} f_0^n}_{\text{not yet diffused (clean)}} +\underbrace{(\sum_{m<k} d_m)\epsilon}_{\text{Gaussian noise from past stages}}.$$

For the usual diffusion process, the SNR is:

\begin{equation}
\text{SNR}^{(t)}
\;:=\;
\frac{\mathbb{E}\!\left[ \left\lVert \sqrt{\bar{\alpha}_t}\,x_0 \right\rVert^2 \right]}
     {\mathbb{E}\!\left[ \left\lVert \sqrt{1-\bar{\alpha}_t}\,\epsilon \right\rVert^2 \right]}
\;=\;
\frac{\bar{\alpha}_t \, \mathrm{Var}(x_0)}{(1-\bar{\alpha}_t)\,\mathrm{Var}(\epsilon)} = \frac{\bar{\alpha}_t}{1-\bar{\alpha}_t}.
\label{eq:SNR_base}
\end{equation}

assuming that $x_0$ is normalized with unit variance.
\subsection{Property 1.}

For each component of our staged diffusion process:
\[
\text{SNR}^{(t)}_{k} \;:=\; 
\frac{\mathbb{E}\, \lvert \sqrt{\bar{\alpha}^*_t}\, f^k_0 \rvert^2}
     {\mathbb{E}\, \lvert \sqrt{d_k(1-\bar{\alpha}^*_t)}\, \varepsilon \rvert^2}
\;=\; \frac{\bar{\alpha^*}_t d_k}{d_k (1-\bar{\alpha^*}_t)} .
\]

\noindent
The $\text{SNR}^{(t)}_{k}$ is strictly decreasing in $t$.
$$
\text{SNR}^{(t)}_{k} 
=\frac{\bar{\alpha}^*_t}{1-\bar{\alpha}^*_t},
$$
note that the SNR equation is quite similar to Eq. \refeq{eq:SNR_base}, with a different sequence of $\{\bar{\alpha}^*\}_{t=1}^\tau$, therefore we know that within each stage the SNR decays smoothly in a known way.

\subsection{Property 2}

At any global step, with component index k and diffusion step t, the total SNR for the observable variable $z^k_t$ is:

\begin{align}
TSNR &= 
\frac{\mathbb{E}\!\left[ \left\lvert f_t^k + \sum_{n>k} f_0^n \right\rvert^2 \right]}
     {\mathbb{E}\!\left[ \left\lvert \sqrt{d_k(1-\bar{\alpha}^*_t)} + \sum_{m<k} d_m \,\epsilon \right\rvert^2 \right]}\\
TSNR &= 
\frac{d_k \bar{\alpha}^*_t + \sum_{n>k} d_n}
     {-\,d_k \bar{\alpha}^*_t + \sum_{m \leq k} d_m}.
\end{align}

and the TSNR at the end of each stage is simply:

\begin{equation}
    TSNR =\frac{\sum_{n>k} d_n}
     {\sum_{m \leq k} d_m} \quad \text{for }t=\tau.
\end{equation}

given that $\alpha_t$ is designed to converge to $0$.
Since $d_k$ is strictly increasing and monotone, we first can guarantee that TSNR is monotone, and that it tends to $0$ when $t\rightarrow \tau$ and $k=K$

An upper bound on the TSNR can be defined, by assuming the edge case where each $d$ assumes the maximum possible value: $d_1 = d_2 =... = d_K = d^*$. Noting that $d^*=1/k$, the resulting upper bound on the TSNR, evaluated at the end of each stage, is given by:

\begin{equation}
    TSNR^+(k) = \frac{K-k}{k}
    \label{eq:TSNR_linear}
\end{equation}

This bound exhibits an inverse dependence on the component number $k$, leading to a monotonically decreasing, hyperbolic decay as the number of stages increases.

\section{Ablation Studies}
\label{sec:experiments}

\subsubsection*{Synthetic Dataset}

This synthetic dataset contains 10000 samples of a signal composed with three different frequencies, with 1000 time-steps each. The goal of this synthetic dataset is to evaluate how the forward diffusion process decomposes the signal, and how the different signals are recuperated during the sampling process.

\begin{algorithm}[h]
\caption{Synthetic Dataset Generation}
\begin{algorithmic}[1]
\STATE \textbf{Input:} $N \in \mathbb{N}$ (number of samples)
\STATE \textbf{Output:} $\{ y^i \}_{i=1}^{N}$ (synthetic time series)
\STATE $f_s \gets 1000$  \COMMENT{Sampling frequency}
\STATE $t \gets \{0, \frac{1}{f_s}, \frac{2}{f_s}, \dots, 1\}$  \COMMENT{Time grid}

\STATE \FOR{$i = 1$ to $N$}
    \STATE $\gamma_{f1}^{(i)} \sim \text{Gamma}(\alpha=3, \beta=5)$
    \STATE $\gamma_{f2}^{(i)} \sim \text{Gamma}(\alpha=4, \beta=5)$
    \STATE $\gamma_{f3}^{(i)} \sim \text{Gamma}(\alpha=5, \beta=5)$
     \STATE $\left( \gamma_{f1}^{(i)}, \gamma_{f2}^{(i)}, \gamma_{f3}^{(i)} \right) \gets \text{sort} \left( \gamma_{f1}^{(i)}, \gamma_{f2}^{(i)}, \gamma_{f3}^{(i)} \right)$
\ENDFOR
\STATE
\FOR{$i = 1$ to $N$}
    \STATE $a_2^{(i)} \sim \text{Uniform}(0.1,1)$
    \STATE $a_3^{(i)} \sim \text{Uniform}(0.1,1)$
\ENDFOR

\FOR{$i = 1$ to $N$}
    \STATE $y_1^i(t) \gets \sin(2\pi \gamma_{f1}^{(i)} t)$
    \STATE $y_2^i(t) \gets a_2^{(i)} \sin(2\pi \gamma_{f2}^{(i)} t)$
    \STATE $y_3^i(t) \gets a_3^{(i)} \sin(2\pi \gamma_{f3}^{(i)} t)$
\ENDFOR

\STATE $y = y_1 +y_2 + y_3$
\end{algorithmic}
\end{algorithm}

\begin{table}[h]
\caption{Synthetic data set details.}
\label{tab:synth_details}
\vskip 0.15in
\begin{center}
\begin{small}
\begin{sc}
\begin{tabular}{lc}
\toprule
Details / Forecast task & \\ 
\midrule
Train size & 9000            \\ 
Test size                            & 1000               \\ 
Training batch                                & 264            \\ 
Sample length                               & 1000                 \\ 
Sample features                             & 1                   \\ 
Conditional values                          & 700                  \\ 
Target values                             & 300                 \\ 
Max Components & 3 \\
\bottomrule
\end{tabular}
\end{sc}
\end{small}
\end{center}
\end{table}

\subsection{Components Ablation}
\label{sec:ablation}
In this section we evaluate the component number. This hyperparameter can be either fixed or variable. In the fixed setting, the number of components is fixed and given as a hyperparameter, in the variable setting, the hyperparameter is only an upper bound. Knowing that the ground truth value of components for this dataset is 3, we evaluate if the diffusion model with the appropriate number of components obtains better results. 
All experiments used trained using the same architecture and the same number of total training steps, with the best model in the validation dataset used for testing. 
\begin{table}[h]
\caption{Forecasting results on synth1 comparing two approaches with varying components ($n$). Ground truth: $n=3$.}
\label{tab:synth_forecast_results}
\vskip 0.15in
\begin{center}
\begin{small}
\begin{sc}
\begin{tabular}{lcccccccc}
\toprule
& \multicolumn{4}{c}{FFT} & \multicolumn{4}{c}{Wavelet} \\
\cmidrule(lr){2-5} \cmidrule(lr){6-9}
Components ($n$) & MSE $\downarrow$ & MAPE $\downarrow$ & MAE $\downarrow$ & CRPS $\downarrow$ & MSE $\downarrow$ & MAPE $\downarrow$ & MAE $\downarrow$ & CRPS $\downarrow$ \\
\midrule
5 & 0.120 & 2.809 & 0.233 & 0.064 & 0.084 & 2.344 & 0.191 & 0.061 \\
4 & 0.189 & 4.147 & 0.307 & 0.083 & 0.082 & 2.464 & 0.186 & 0.052 \\
\textbf{3} (True Value) & \textbf{0.078} & \textbf{1.794} & \textbf{0.177} & 0.045 &  \textbf{0.066} & \textbf{1.382} & \textbf{0.170} & \textbf{0.047} \\
2 & 0.119 & 3.640 & 0.234 & 0.060 & 0.076 & 2.091 & 0.183 & 0.049 \\
1 & 0.098 & 2.056 & 0.185 & \textbf{0.043} & 0.072 & 2.181 & 0.176 & \textbf{0.047} \\
\midrule
\multicolumn{9}{l}{ Baseline ($n=0$): MSE=0.095, MAPE=2.620, MAE=0.204, CRPS=0.055} \\
\bottomrule
\end{tabular}
\end{sc}
\end{small}
\end{center}
\end{table}

\begin{figure}[!h]
    \centering
    \begin{subfigure}{0.49\textwidth}
        \centering
        \includegraphics[width=\linewidth]{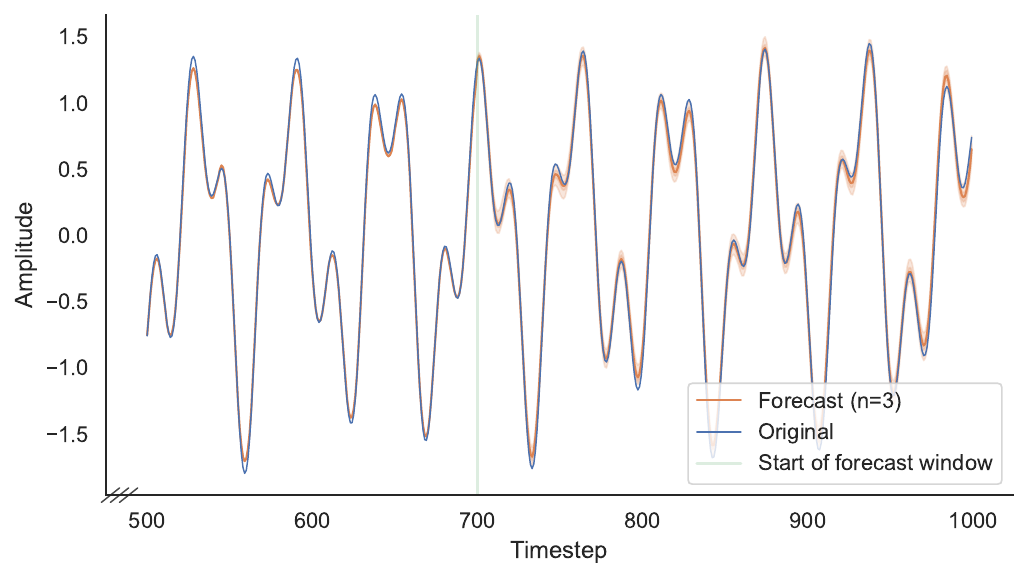}
        \caption{1 component}
    \end{subfigure}
    \begin{subfigure}{0.49\textwidth}
        \centering
        \includegraphics[width=\linewidth]{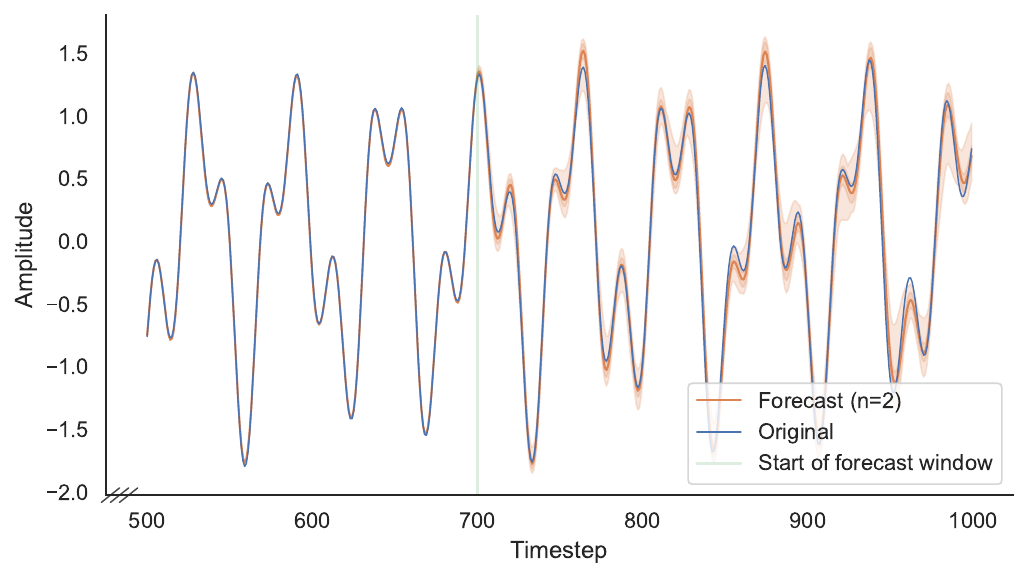}
        \caption{2 components}
    \end{subfigure}
    \begin{subfigure}{0.49\textwidth}
        \centering
        \includegraphics[width=\linewidth]{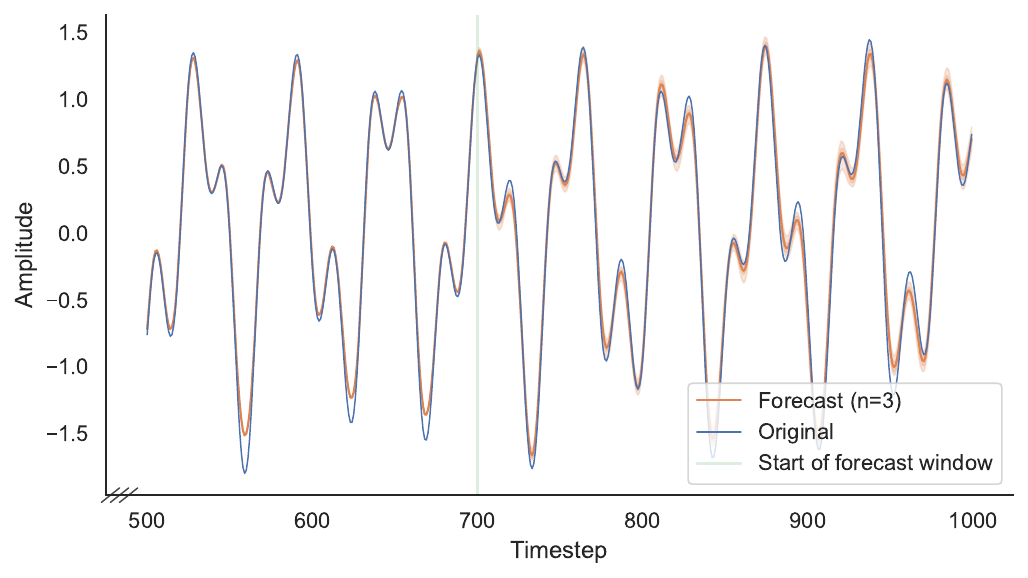}
        \caption{3 components}
    \end{subfigure}
    \begin{subfigure}{0.49\textwidth}
        \centering
        \includegraphics[width=\linewidth]{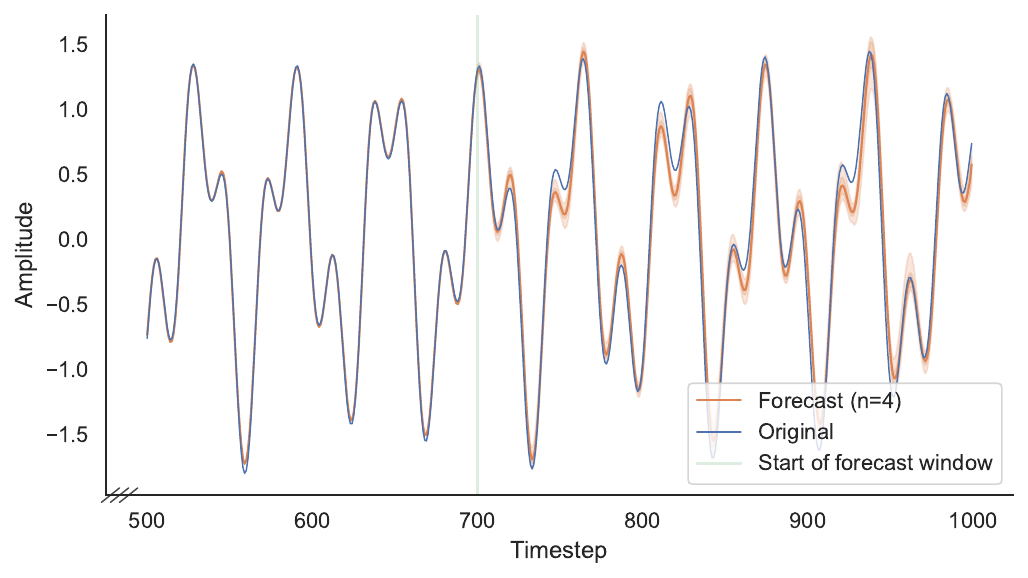}
        \caption{4 components}
    \end{subfigure}
\label{plots}
\caption{forecasting comparison for a 300 steps window length, of choosing a different number of components.}
\end{figure}

Table~\ref{tab:synth_forecast_results} shows forecasting performance as a function of the number of components
$n$, with all other hyperparameters fixed. The model with 
$n=3$ components achieves the best performance across MSE, MAPE, and MAE. Among the other models, $n=1$ performs best, suggesting that a single component can capture meaningful structure by separating the diffusion process into two distinct components. We also observe that smaller values of $n$ consistently yield lower CRPS scores.

While both FFT and Wavelet decomposition obtained the best results using the ground truth component value, the Wavelet decomposition seems to be more robust, or less sensitive to this hyperparameter, as all values achieve results better than the baseline. 

The baseline, which here represents the usual diffusion process, provides a slightly worst forecast compared with the correct number of components,  especially in MAPE.

From this we can conclude that the component-based diffusion has a significant effect on the diffusion process, and that choosing the correct hyperparameter $k$ has a positive effect on the final result.

\subsection{Scheduler parameters}
\begin{figure}[h]
  \centering
  \includegraphics[]{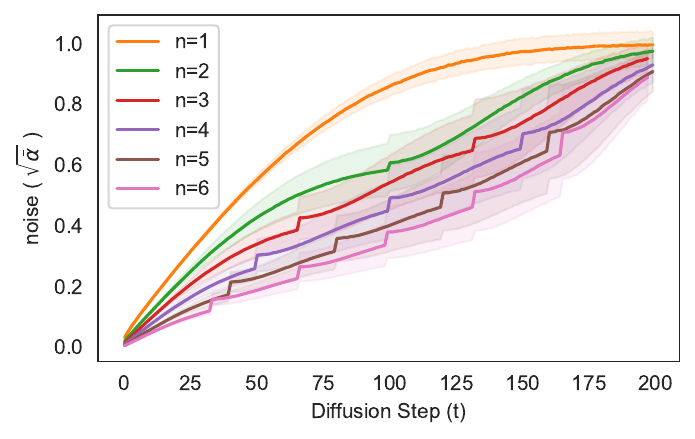}
  \caption{Evolution of the noise scheduler with different number of components. As the number of components increase, the smoothness of the diffusion process decreases, for fixed scheduler parameters.}
  \label{fig:example_noise}
\end{figure}

This ablation study analyzes how the scheduler is affected. For a fixed T, as the number of components increase, the number of diffusion steps per component decrease. This leads to a less smooth, not continuous forward diffusion process, with gaps when transitioning between components.
Furthermore the last component is, by definition, the one with smaller amplitude, and therefore, if more components than necessary are added, the final components of the reverse diffusion have a very limited impact. In Fig. \ref{fig:example_noise} we can observe this effect using a toy example, with the average noising process of 1000 training examples, as we add components and fix the scheduler, using ($\beta_0 = 0.01, \beta_T = 0.05$). The more components are added, the less smooth is the forward diffusion process, and also, for $n\geq 3$, the forward diffusion process does not reach the usual diffusion standard deviation of 1. 

To solve this, we identify the parameter space that is able to satisfy our requirements of smoothness, and guarantee that the final step of the diffusion process has $z_T^K \sim N(0,1)$.

Given a linear noise schedule in a diffusion process, we want it to satisfy a global bound on cumulative noise $\bar{\alpha}_\tau = \prod^\tau_{t=0} \alpha$. Let $\tau$ denote the total number of diffusion steps for a given component, and let $\beta_0$ and $\beta_\tau$ be the starting and ending noise parameters, respectively. A linear scheduler can be trivially defined as:
\begin{equation*}
    \beta_t = \beta_0 + \frac{t}{\tau} (\beta_\tau - \beta_0), \quad \text{for } t = 0, 1, \ldots, \tau.
\end{equation*}

Our goal is to enforce the constraint that $\bar{\alpha}_\tau$ is bellow a threshold $q$:
\begin{equation}
    \prod_{t=0}^{\tau} (1 - \beta_t) \leq q,
    \label{eq:prod-constraint}
\end{equation}
where typically $q \in ]0, 0.1]$, and considering the constraints $b_\tau > b_0$ and $b_\tau, b_0 > 0$.

\begin{figure}[h]
  \centering
  \includegraphics[width=\linewidth]{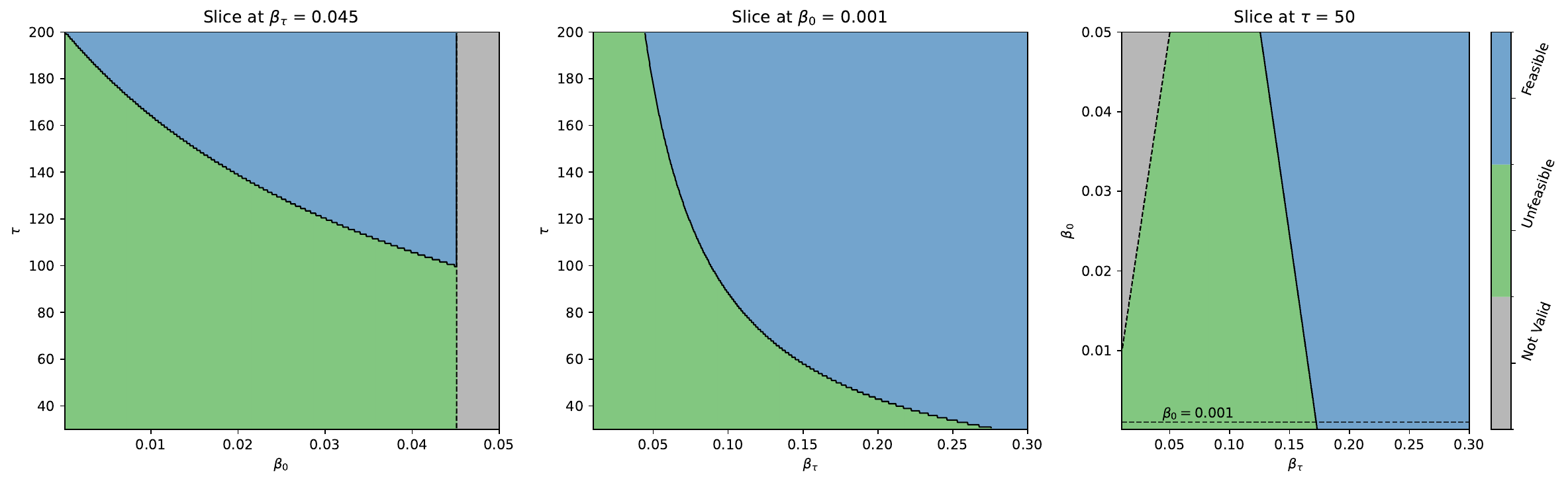}
  \caption{Exploration of the feasibility region for $T \in [30,200], \beta_\tau \in [0.01,0.3], \beta_0 \in [0.0001, 0.05]$. The gray area indicates where  $b_\tau > b_0$,with the green area indicating the feasibility space.}
  \label{fig:ineq}
\end{figure}

The solution can be solved numerically for $\tau,\beta_\tau$ and $\beta_0$ as seen on Fig.\ref{fig:ineq}. Note that these values are independent of the data. Since, for the usual $T=200$, $\tau$ depends on the number of components, and choosing a fixed $\beta_0 =0.001$ we can approximate the minimum value of $\beta_\tau$ that follows the constraint.

\begin{figure}[!h]
  \centering
  \includegraphics[]{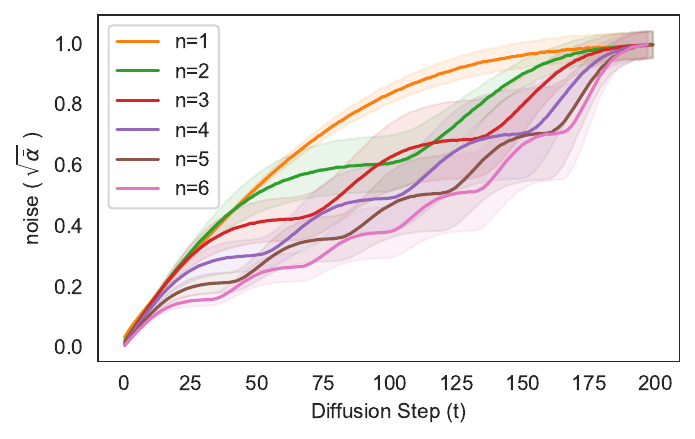}
  \caption{Evolution of the noise scheduler, with tuned $\beta_\tau$. By tuning $\beta_\tau$, the noise scheduler is able to reach a value close to zero when $t=\tau$, regardless of the number of components, or the average amplitude of each component. 
}
  \label{fig:example_noise2}
\end{figure}

By choosing appropriate scheduler parameters, a smoother, more continuous growth of the noise added to the sample is obtained, as can clearly be see in Fig. \ref{fig:example_noise2}. Also, this guarantees that $\sqrt{\bar{\alpha_t}}$ converges to 1 at the end of the diffusion process. 

\newpage

\section{Computational Cost}
\label{app:cost}

We analyze both the theoretical and empirical computational cost of our decomposition-based diffusion model, using the FFT decomposition. Although similar in the process, the algorithmic cost of the wavelet decomposition is lower, since the computational cost of the Wavelet transform is well known to be $\mathcal{O}(N)$ \cite{mallat,fossgaard1999fast}.

\subsection{Theoretical Cost for the FFT}
The FFT decomposition function requires performing a Fast Fourier Transform (FFT) over the sample batch, selecting the top-$k$ frequencies, and then applying an inverse FFT (IFFT) to reconstruct the chosen components. The sorting step is negligible, since the number of selected components $k$ is much smaller than the number of time steps $N$.  

Overall, the computational complexity of our method using the FFT is:
\begin{equation}
    \mathcal{O}(2N \log N).
\end{equation}

In practice, the additional cost is primarily observed during training, where the decomposition is performed for each batch. During inference, the decomposition is executed only once every $T$ diffusion steps, so the overhead remains very limited.

\subsection{Algorithmic}

We summarize the decomposition procedure in Algorithm~\ref{alg:fft-decomp}, for simplicity, for just one channel, since the extension is trivial.

\begin{algorithm}[h]
\caption{FFT-based Decomposition}
\label{alg:fft-decomp}
\begin{algorithmic}[1]
\STATE Batch of time-series $ X = \{x_0\}_{i=1}^B$, 
\STATE $\tilde{F} \gets \mathbf{0}_{B \times k \times N}$
\STATE $F \gets \text{FFT}(X)$
\STATE $I \gets \text{indices of top-$k$ frequencies in each row of } |F|$
\STATE $\tilde{F} \gets F[:, I]$ \COMMENT{keep only top-$k$ frequencies per sequence, in separate components}
\STATE $D \gets \text{IFFT}(\tilde{F})$ 
\STATE Sort($D$)

\end{algorithmic}
\end{algorithm}


\subsection{Empirical Cost}
We measure the practical overhead in terms of average increase in training time per epoch across multiple datasets and for both the FFT and Wavelet decomposition. The results are summarized in Table~\ref{tab:cost}. We observe an average overhead of approximately 7.5\% per epoch, which is modest given the accuracy improvements.

\begin{table}[h]
    \centering
    \caption{Increase in training time per epoch due to FFT-based decomposition.}
    \label{tab:cost}
    \begin{tabular}{lccccccc}
        \toprule
        \textbf{datasets} & Electricity & ETTm1 & Mujoco & PTB-XL & Synth1 & Temperature & Solar \\
        \midrule
         & 11.79\% & 1.19\% & 11.55\% & 4.19\% & 12.10\% & 4.25\% & 4.11\% \\
        \bottomrule
    \end{tabular}
\end{table}






\section{Forecast Figures}
\label{sec:figures}

\begin{figure}[!th]
    \centering
    \begin{subfigure}{0.30\textwidth}
        \centering
        \includegraphics[width=\linewidth]{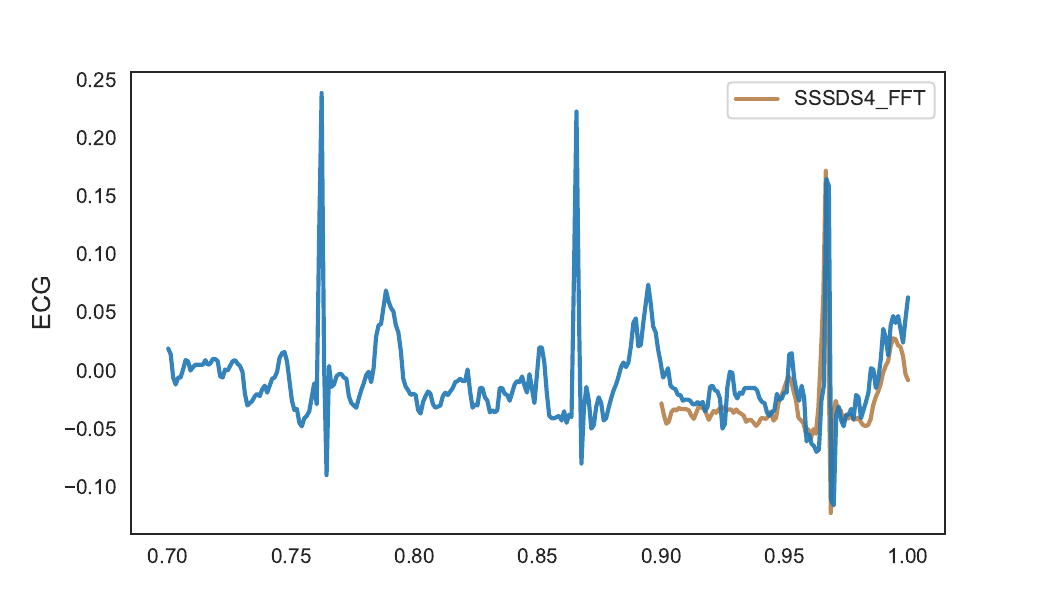}
        \caption{SSSD + FFT}
    \end{subfigure}
    \begin{subfigure}{0.30\textwidth}
        \centering
        \includegraphics[width=\linewidth]{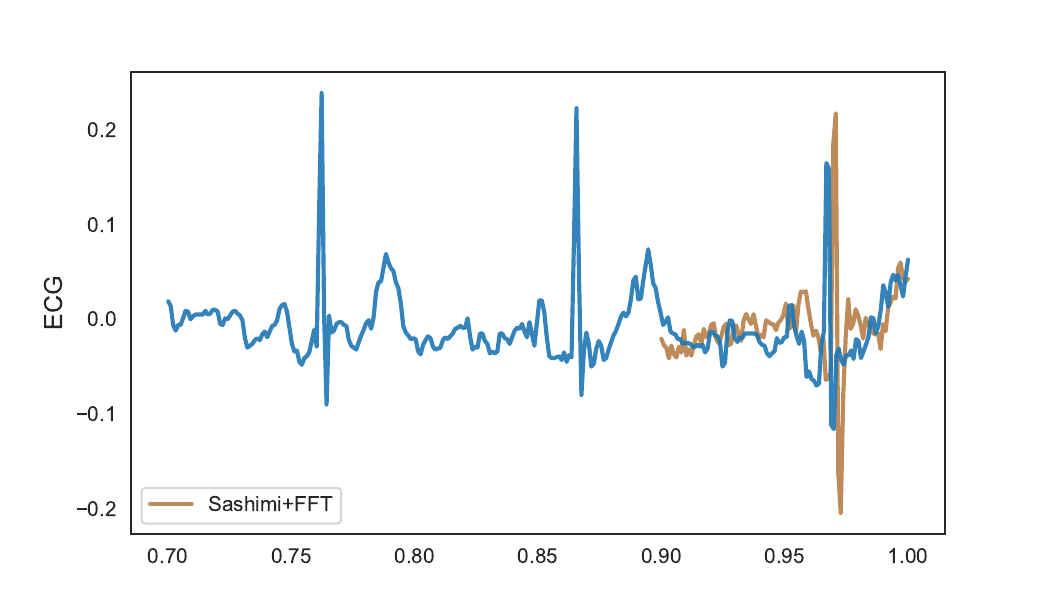}
        \caption{Sashimi + FFT}
    \end{subfigure}
    \begin{subfigure}{0.30\textwidth}
        \centering
        \includegraphics[width=\linewidth]{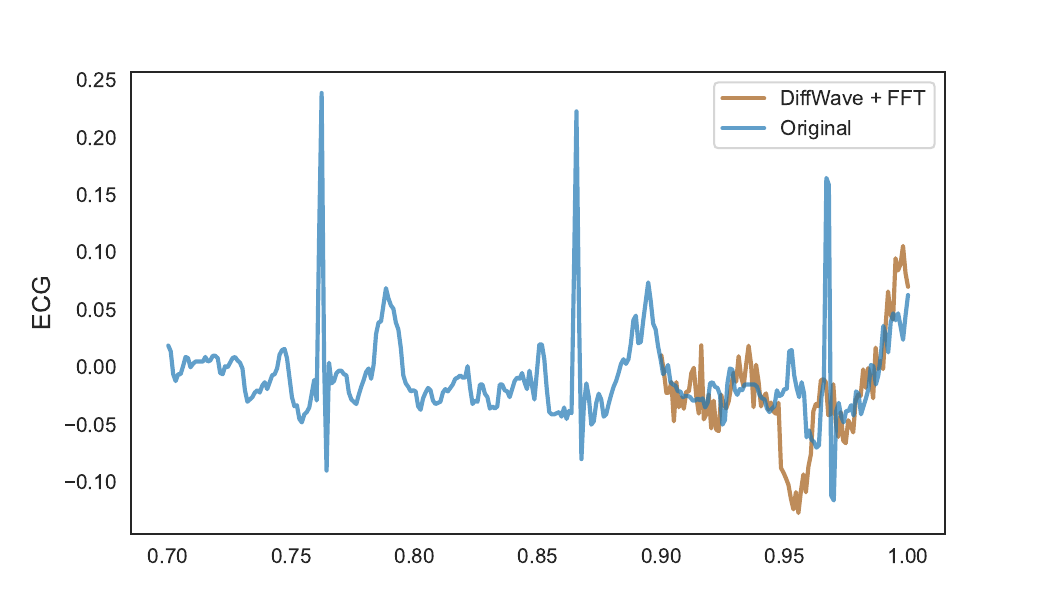}
        \caption{DiffWave + FFT}
    \end{subfigure}
    \begin{subfigure}{0.30\textwidth}
        \centering
        \includegraphics[width=\linewidth]{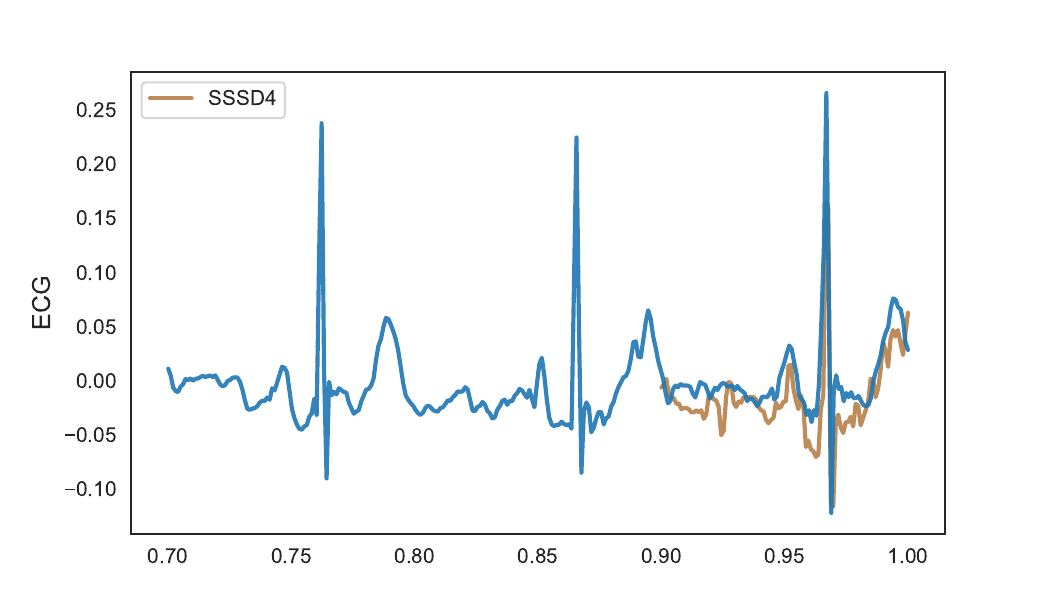}
        \caption{SSSD}
    \end{subfigure}
    \begin{subfigure}{0.30\textwidth}
        \centering
        \includegraphics[width=\linewidth]{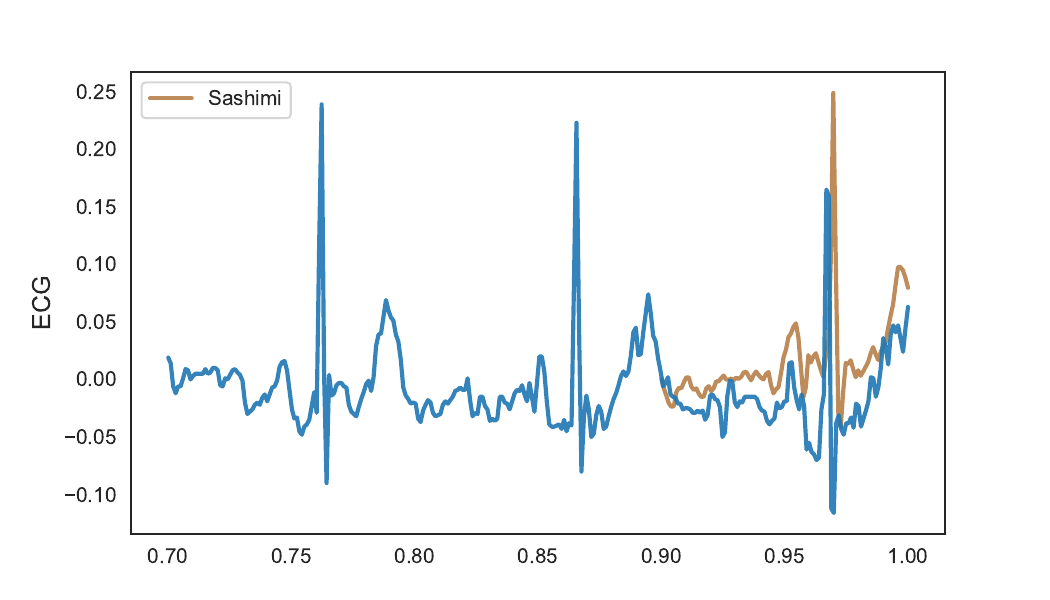}
        \caption{Sashimi}
    \end{subfigure}
    \begin{subfigure}{0.30\textwidth}
        \centering
        \includegraphics[width=\linewidth]{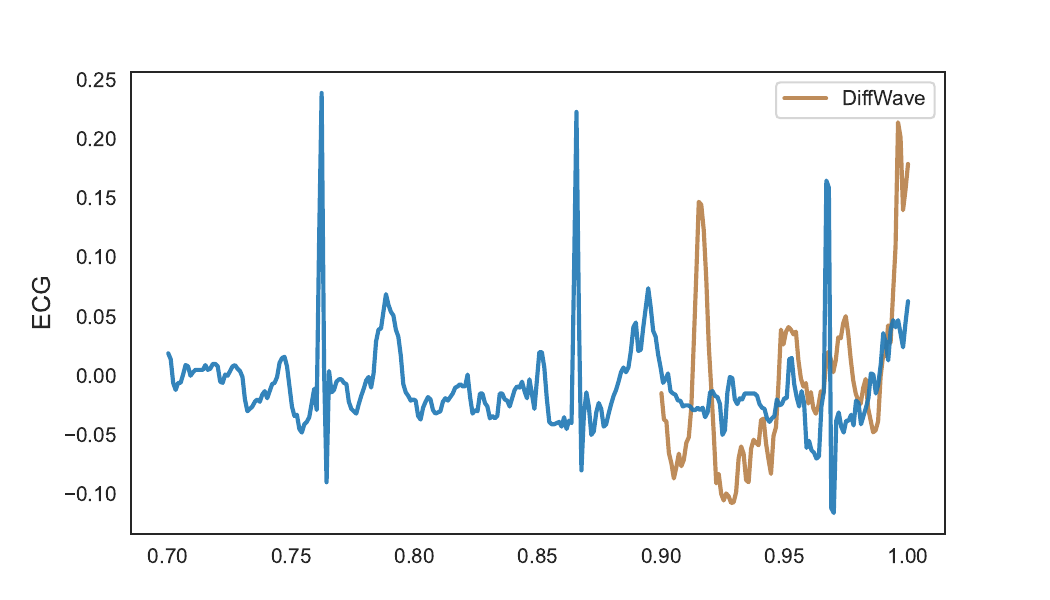}
        \caption{DiffWave}
    \end{subfigure}
\caption{One generation for a random sample of the PTB-XL dataset, for three different models with and without FFT decomposition.}
\label{example_forecast}
\end{figure}

\end{document}